\pdfoutput=1

\documentclass[10pt,twocolumn,letterpaper]{article}

\usepackage{multirow}
\usepackage{fontawesome5}
\usepackage[pagenumbers]{cvpr} 

\definecolor{cvprblue}{rgb}{0.21,0.49,0.74}
\usepackage[pagebackref,breaklinks,colorlinks,allcolors=cvprblue]{hyperref}

\def\confName{CVPR}
\def\confYear{2027}

\title{TANGO: Real-World Unregistered RGB-Guided Hyperspectral Super-Resolution with Two-Camera Acquisitions}

\author{
Matteo Kolyszko\quad
Marco Buzzelli\quad
Simone Bianco\quad
Raimondo Schettini\\
University of Milano-Bicocca\\
{\tt\small \{\href{mailto:matteo.kolyszko@unimib.it}{\textcolor{black}{matteo.kolyszko}}, \href{mailto:marco.buzzelli@unimib.it}{\textcolor{black}{marco.buzzelli}}, \href{mailto:simone.bianco@unimib.it}{\textcolor{black}{simone.bianco}}, \href{mailto:raimondo.schettini@unimib.it}{\textcolor{black}{raimondo.schettini}}\}@unimib.it} 
}

\begin{document}
\maketitle
\begin{abstract}
Unregistered hyperspectral image super-resolution aims to recover a high-resolution HSI from a low-resolution hyperspectral observation and an unregistered high-resolution RGB reference. Although recent methods have begun to address the geometric misalignment between the two modalities, their evaluation is still constrained by the lack of a large-scale public benchmark based on real two-camera RGB--HSI acquisitions with a high-resolution HSI reference. To address this gap, we introduce DUET, the first large-scale public dataset for unregistered RGB-guided hyperspectral super-resolution, comprising 403 indoor and outdoor scenes captured with two separate RGB and hyperspectral cameras. We further propose TANGO (\emph{Two-camera AligNment with Gated Occlusions}), which first compensates for the dominant cross-camera displacement through end-to-end global alignment and then refines the remaining inconsistencies with a coarse-to-fine local feature alignment. A learned gating mechanism further attenuates the RGB guidance where the two views cannot be consistently matched. Experimental results show that TANGO achieves state-of-the-art performance on both DUET and REAL across all evaluated scale factors, outperforming the strongest competitor on DUET by up to 2.29\,dB in PSNR while requiring 2.5$\times$ fewer FLOPs. The code and dataset will be available at {\url{https://github.com/FagiolottoConSombrero/TANGO-DUET}}
\end{abstract}    
\section{Introduction}
\label{sec:introduction}

\begin{figure}[t]
    \centering
    \includegraphics[width=\columnwidth]{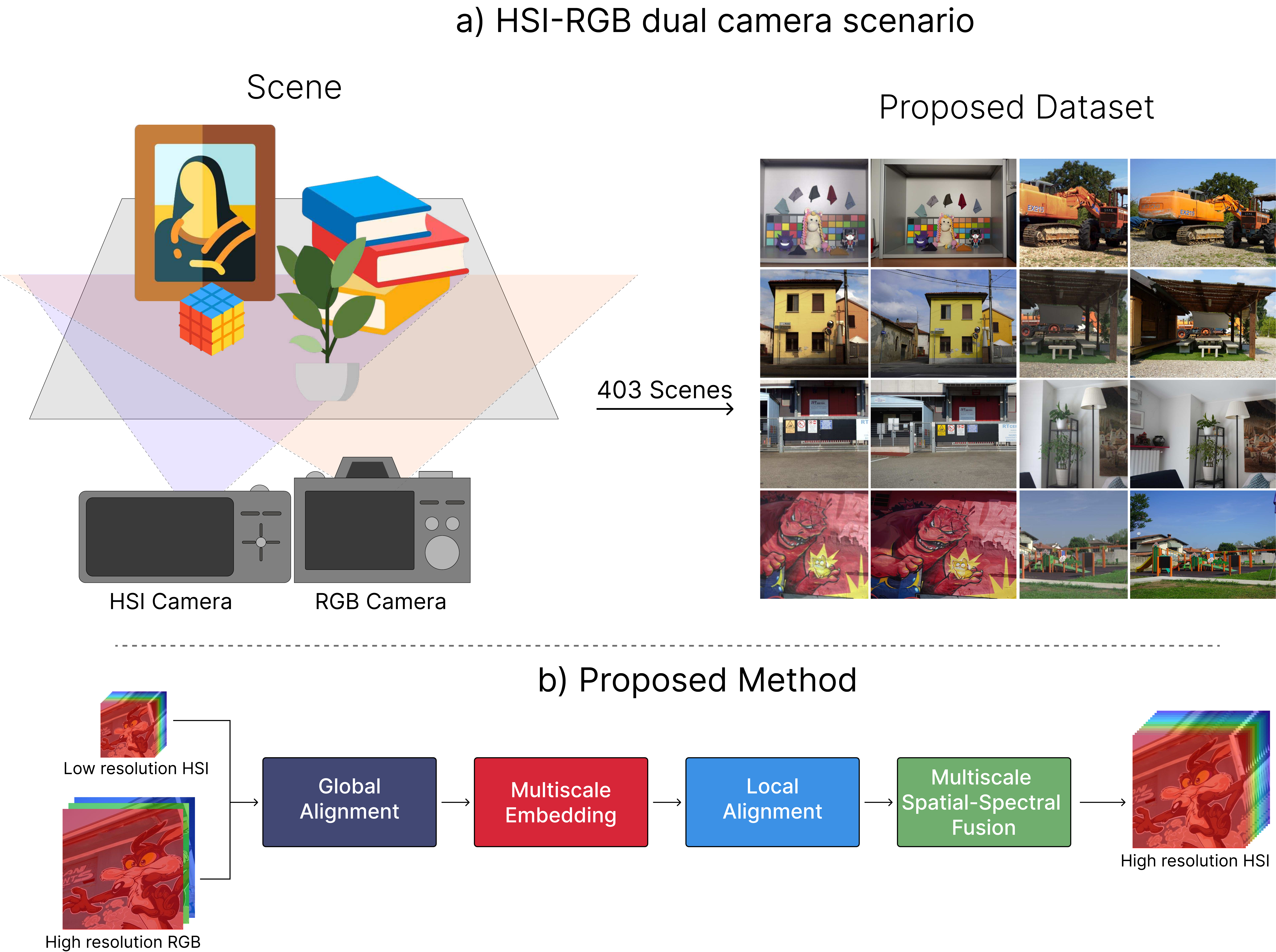}
    \caption{Overview of the proposed framework. (a) We illustrate the dual-camera acquisition setup used to collect DUET, where an RGB camera and a hyperspectral camera are mounted side by side to capture 403 real HSI--RGB scene pairs under natural cross-camera misalignment. (b) We propose TANGO, a model for unregistered RGB-guided hyperspectral super-resolution that reconstructs a high-resolution HSI from a low-resolution HSI and an unregistered high-resolution RGB image.}
    \label{fig:overview}
\end{figure}

Hyperspectral imaging provides dense spectral measurements across tens or
hundreds of contiguous wavelength bands, enabling fine-grained analysis of
materials and scene properties beyond the capabilities of conventional RGB
imaging. This rich spectral representation has made hyperspectral imaging
valuable in a wide range of applications~\cite{hajaj2024review, ram2024systematic, lu2014medical, antony2024hyperspectral, lopez2026hyperspectral, vasile2024space}. However, hyperspectral acquisition
is inherently affected by a trade-off between spatial resolution, spectral
resolution, and signal-to-noise ratio. Consequently, hyperspectral images often
provide detailed spectral information at the cost of limited spatial detail,
making hyperspectral image super-resolution (HSI-SR) an important problem. A first line of research addresses this limitation by relying exclusively on the low-resolution HSI, exploiting the intrinsic spatial and spectral correlations of the hyperspectral cube to infer the missing high-frequency spatial information through increasingly effective spectral-spatial processing, attention mechanisms, and efficient feature extraction~\cite{FPB2020, EssaFormer, kolyszko2026decoupling}.
Nevertheless, single-image methods remain fundamentally constrained by the
spatial information available in the low-resolution observation itself. To overcome this limitation, RGB-guided HSI super-resolution introduces an
auxiliary high-resolution RGB image that provides complementary high-frequency
spatial information. Several methods have demonstrated that combining the
spectral content of the HSI with the spatial detail of an RGB reference can
substantially improve reconstruction quality
~\cite{Fu2019HSISR, liu2025low, kolyszko2026multiscale}. However,
most RGB-guided methods assume that the RGB and hyperspectral observations are
spatially registered before fusion.

This assumption becomes problematic when the two modalities are acquired with
independent cameras. Differences in viewpoint, focal length, sensor geometry,
and scene depth introduce global displacement, perspective changes,
depth-dependent parallax, local geometric inconsistencies, and occlusions.
Under these conditions, directly injecting RGB features into the hyperspectral
reconstruction may transfer spatial information to incorrect locations and
introduce reconstruction artifacts.

Recent methods have therefore started to address unregistered RGB-guided HSI super-resolution by explicitly modeling the spatial discrepancies between the two modalities through registration, correspondence estimation, or deformable feature aggregation integrated into the multimodal reconstruction process~\cite{fu2020simultaneous,Zheng2022NonRegSRNet,lai2024hyperspectral,Dong_Xu_Qu_Hou_2024,zhang2025unaligned,zhang2026enhancing}. However, existing benchmarks only partially reproduce this setting. REAL~\cite{lai2024hyperspectral} is acquired with two hyperspectral cameras, and its RGB guidance is synthesized from one of the hyperspectral observations, so cross-sensor discrepancies are not represented. A recent work~\cite{jouni2026lightweight} uses a real RGB--HSI dual-camera rig, but evaluates on a single static indoor sequence of 34 frames, with quantitative results limited to a reduced-resolution protocol at $\times4$. As a result, no existing benchmark combines real two-camera RGB--HSI acquisitions with the scale, scene diversity, and range of scale factors required to train and evaluate learning-based methods.

To address this gap, we introduce \textbf{DUET}, the first large-scale public dataset specifically designed for unregistered RGB-guided HSI super-resolution using real two-camera acquisitions. DUET contains 403 scenes (30 indoor and 373 outdoor) independently captured with a Sony $\alpha$6100 RGB camera and a Specim IQ hyperspectral camera mounted side by side, and provides a benchmark protocol at $\times4$, $\times8$, and $\times16$. The dataset therefore preserves the viewpoint differences, depth-dependent parallax, occlusions, and cross-sensor appearance variations naturally produced by independent imaging systems.

Building on this setting, we further propose \textbf{TANGO}
(\emph{Two-camera AligNment with Gated Occlusions}), a reconstruction framework
designed to exploit RGB guidance under real cross-camera misalignment. TANGO combines global alignment with coarse-to-fine local feature refinement and a learned gating mechanism. The global stage compensates for the dominant
inter-camera displacement, while the local alignment modules progressively
refine residual correspondences at multiple feature scales. The gating mechanism, motivated by occlusions and inaccurate correspondences, further attenuates unreliable RGB information. An overview of the considered
dual-camera acquisition setting, the proposed DUET dataset, and the TANGO
reconstruction pipeline is shown in Fig.~\ref{fig:overview}.

Our contributions are summarized as follows:
\begin{itemize}
    \item We introduce DUET, the first large-scale public benchmark for unregistered RGB-guided HSI super-resolution based on real acquisitions from separate RGB and hyperspectral cameras, comprising 403 indoor and outdoor scenes.

    \item We propose TANGO, which combines global alignment, multiscale local refinement, and learned gating to robustly exploit unregistered RGB guidance.

    \item We demonstrate state-of-the-art performance on both DUET and
    REAL datasets across all evaluated scale factors.
\end{itemize}

\section{Related Work}
\label{sec:related_work}

\noindent{\textbf{RGB-guided HSI super-resolution.}}
RGB-guided HSI super-resolution exploits a high-resolution RGB or multispectral
observation to recover spatial details that are missing from a low-resolution
hyperspectral image. Early approaches mainly formulated the problem through
model-based optimization, relying on sparse representations, matrix
factorization, tensor decomposition, and other handcrafted priors
~\cite{Akhtar2014ECCV,Akhtar2015CVPR,Akhtar2016ECCV,Kawakami2011CVPR,
Dong2016TIP,Li2018TIP}. With the development of deep learning, these
formulations were progressively replaced by data-driven fusion architectures
that jointly exploit the spatial information of the guidance image and the
spectral information of the HSI
~\cite{Fu2019HSISR,Zhou2019Fusion,Hu2022TNNLS,Qu2022MutualDirichletNet,
Guo2023CVPR,Wu2023ICCV,Zhang2024DPL, kolyszko2026multiscale}.More recent approaches have increasingly adopted attention and
Transformer-based mechanisms to improve cross-modal interaction and capture
long-range spatial-spectral dependencies
~\cite{liu2025low,Zhang2025Dual}. Despite their strong reconstruction performance, however, these RGB-guided
approaches generally assume that the two modalities are spatially registered
before fusion. This assumption is difficult to satisfy in practical acquisition
systems, where RGB and hyperspectral observations are often captured by
different sensors with distinct optics, resolutions, fields of view, and
viewpoints. 

\noindent{\textbf{Unregistered RGB-guided HSI super-resolution.}}
The requirement for pixel-aligned multimodal observations is difficult to
satisfy in practical systems, where independent cameras introduce viewpoint
changes, scale variations, perspective differences, parallax, and occlusions.
Several works have therefore investigated joint registration and reconstruction
for hyperspectral fusion
~\cite{fu2020simultaneous,nie2020unsupervised,Qu2022MutualDirichletNet,
Zheng2022NonRegSRNet}. Earlier approaches explored explicit geometric
registration or non-rigid transformations before fusion
~\cite{fu2020simultaneous,nie2020unsupervised,Zheng2022NonRegSRNet},
while other formulations sought to integrate registration and reconstruction
more tightly within a unified optimization or learning framework
~\cite{Qu2022MutualDirichletNet,Ying2022Unaligned,Qu2024MoEPNP}.

More recent methods increasingly combine correspondence estimation with deep
multimodal feature fusion
~\cite{lai2024hyperspectral,Dong_Xu_Qu_Hou_2024,
zhang2025unaligned,zhang2026enhancing}. Overall, these methods
reflect a progressive shift from assuming perfectly aligned inputs toward
explicitly modeling geometric inconsistencies and correspondence uncertainty
during reconstruction
~\cite{fu2020simultaneous,nie2020unsupervised,Qu2022MutualDirichletNet,
Zheng2022NonRegSRNet,lai2024hyperspectral,Dong_Xu_Qu_Hou_2024,
zhang2025unaligned,zhang2026enhancing}. 

Despite this progress, the evaluation of unregistered RGB-guided HSI super-resolution remains limited by the available data. REAL~\cite{lai2024hyperspectral} contains hyperspectral observations captured from different viewpoints, while the RGB guidance is synthetically generated from one of the hyperspectral cubes, so the cross-sensor discrepancies of a real RGB--HSI system are not represented. A recent model-based approach~\cite{jouni2026lightweight} acquires native RGB--HSI pairs with a snapshot hyperspectral camera, but is evaluated on a single static indoor sequence of 34 frames, with quantitative results limited to a reduced-resolution protocol at $\times4$. It also modulates the RGB guidance with a pixel-wise reliability map, computed from hand-crafted photometric, structural, and cycle-consistency cues and applied once to blend a Gram--Schmidt fusion with the upsampled HSI. DUET addresses these limitations by providing 403 real paired acquisitions from separate RGB and hyperspectral cameras, while preserving the residual misalignment naturally introduced by the acquisition geometry. Building on this setting, TANGO combines global alignment with coarse-to-fine local correspondence refinement and a learned gating mechanism, explicitly targeting the parallax, local misalignment, and missing correspondences that remain after a global registration stage.
\section{DUET Dataset}
\label{sec:DUET_dataset}

DUET contains 403 scenes captured with a Sony $\alpha$6100~\cite{sonyA6100} RGB camera and a Specim IQ~\cite{specimIQ} hyperspectral camera mounted side by side on a common plate, which results in approximately 15\,cm between their optical centers and produces a depth-dependent parallax between the two views. The acquisition setup and its geometry are described in 
the supplementary material. The Sony camera acquires RGB images at approximately $6000\times4000$ pixels, whereas the Specim IQ provides $512\times512$ cubes with about 204 bands in 397--1004\,nm, from which we retain 31 bands between 400 and 700\,nm at 10\,nm intervals. The dataset includes 30 indoor scenes, recorded under tungsten illumination, and 373 outdoor scenes, captured under natural daylight at different times of day, covering diverse materials, geometries, and depths. Representative pairs are shown in Fig.~\ref{fig:DUET_samples}, and further examples are provided in 
the supplementary material.

\begin{figure*}[t]
    \centering
    \includegraphics[width=\linewidth]{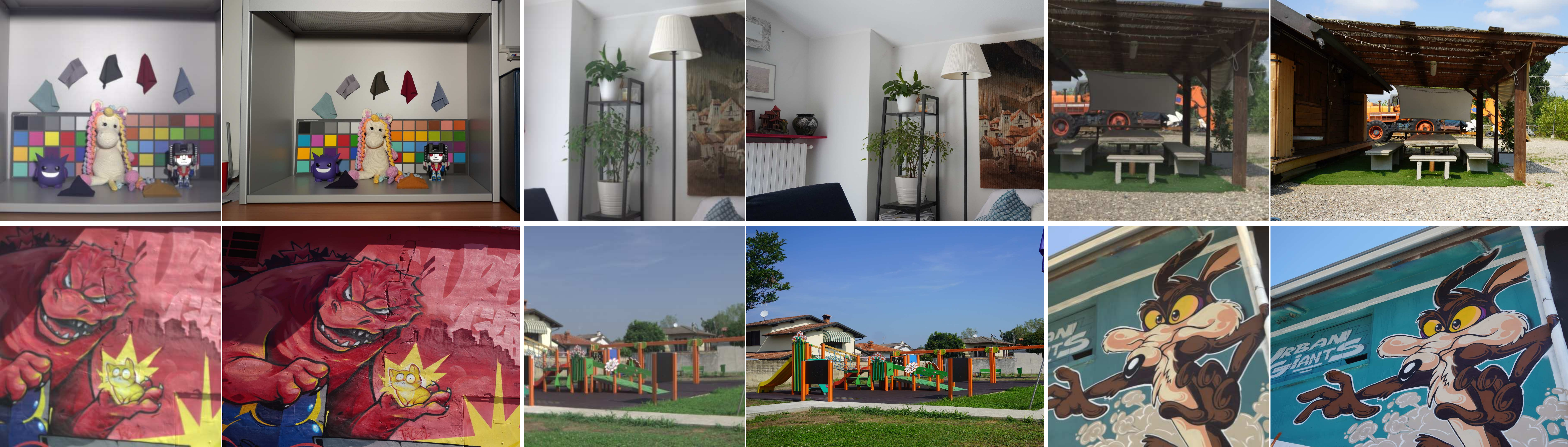}
    \caption{Representative RGB--HSI pairs from DUET. RGB images are shown as the original in-camera JPEGs, with the full Sony field of view and no processing. Hyperspectral cubes are rendered as pseudo-RGB images from the bands at 650, 560, and 470\,nm. The pairs highlight scene diversity and the real viewpoint and field-of-view differences between the two cameras.}
    \label{fig:DUET_samples}
\end{figure*}

\noindent\textbf{Preprocessing.}
Sony RAW images are converted to linear RGB without white balancing, gamma correction, brightness adjustment, or color-space transformation, and the Specim IQ cubes are converted to reflectance using a white reference panel acquired before each scene. Because the two cameras have different fields of view and spatial resolutions, a coarse field-of-view matching is then performed. SIFT features are matched using FLANN~\cite{muja2009fast} and filtered with Lowe's ratio test~\cite{lowe2004distinctive}, and the resulting correspondences are used within RANSAC~\cite{fischler1981random} to estimate a homography between the two views. Rather than fully warping the RGB image, the homography is approximated by an isotropic scaling and translation, fitted with least squares to a regular grid of HSI pixels mapped through the homography. The corresponding square RGB region is then cropped and resized to $512\times512$. This removes the largest field-of-view differences while intentionally preserving residual rotation, perspective distortion, parallax, and occlusions. The procedure is illustrated in 
the supplementary material.

\noindent\textbf{Benchmark Protocol.}
The $512\times512$ hyperspectral cubes are used as ground truth, and low-resolution inputs are generated at $\times4$, $\times8$, and $\times16$ with the Gaussian blur and decimation protocol of~\cite{lai2024hyperspectral}, detailed in 
the supplementary material. The dataset is divided into 327 training, 36 validation, and 40 test scenes. Since the scenes were acquired over ten sessions on different days, with different locations and illumination conditions, both the test and validation scenes are selected per session, taking four test scenes and three to four validation scenes from each session, so that every acquisition condition is represented in all splits.

\noindent\textbf{Misalignment Statistics.}
To quantify the residual misalignment, we estimate a dense bidirectional flow between the pseudo-RGB of each HR-HSI and the corresponding RGB image with RAFT~\cite{teed2020raft}, fit a RANSAC homography to it, and measure the non-planar residual, \ie, the displacement not explained by the homography. For the median scene, the 95th percentile of this residual is 11.1\,px, and it exceeds 4\,px in 72.7\% of the scenes; on average, 9.6\% of the pixels fail the forward--backward consistency check and therefore have no reliable correspondence. DUET thus cannot be registered by a single global transformation, which motivates the dense and local alignment stages of TANGO. Detailed statistics and correspondence maps are provided in 
 the supplementary material.
\section{Proposed Method}
\label{sec:method}

\noindent\textbf{Problem Formulation.}
Given a low-resolution hyperspectral image
$\mathbf{Y}\in\mathbb{R}^{C\times h\times w}$ and an unregistered
high-resolution RGB image
$\mathbf{R}\in\mathbb{R}^{3\times H\times W}$, the goal is to reconstruct a
high-resolution hyperspectral image
$\hat{\mathbf{X}}\in\mathbb{R}^{C\times H\times W}$ in the hyperspectral
camera coordinate system.

\begin{figure*}[t]
    \centering
    \includegraphics[width=\linewidth]{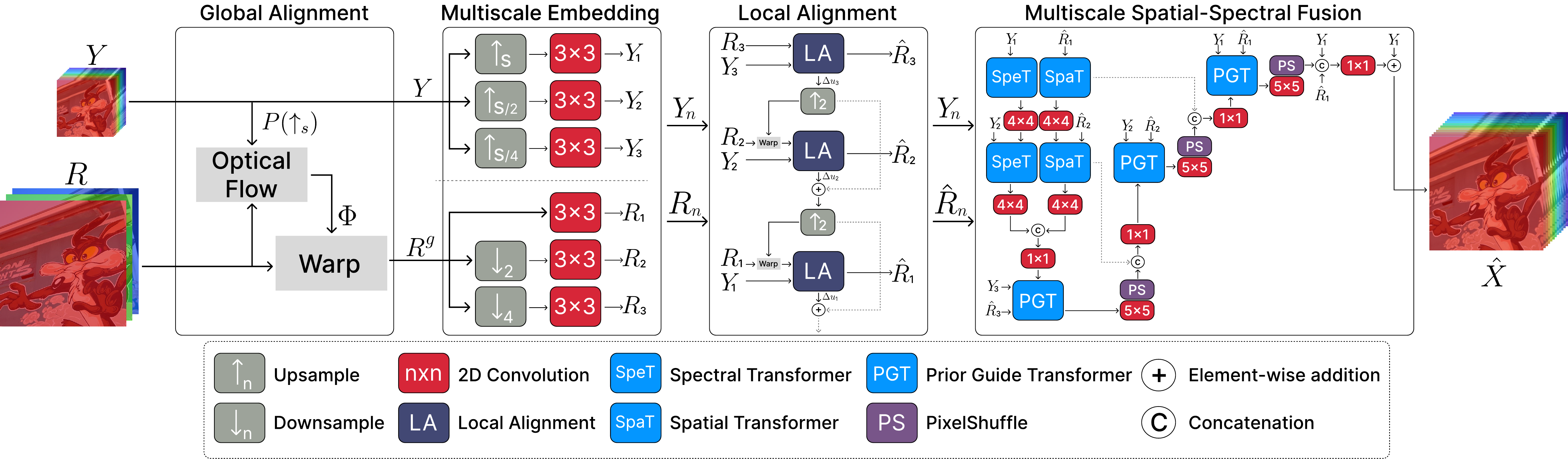}
    \caption{Overview of TANGO. Global Alignment first compensates for the
    dominant displacement between the HSI and RGB views. Multiscale Embedding
    extracts features at three resolutions, which are progressively refined by
    the Local Alignment modules. The resulting aligned and gated RGB features
    are finally combined with the HSI representations through the multiscale
    spatial-spectral fusion backbone to reconstruct $\hat{\mathbf{X}}$.}
    \label{fig:tango_overview}
\end{figure*}

\noindent\textbf{Overview.}
As illustrated in Fig.~\ref{fig:tango_overview}, TANGO addresses this problem
through four stages: Global Alignment, Multiscale Embedding, Local Alignment,
and Multiscale Spatial-Spectral Fusion. The first stage compensates for the
dominant displacement between the two camera views, after which both modalities
are embedded at multiple spatial scales. Residual misalignment is then refined
progressively in feature space, while a learned gate modulates the contribution of the transferred RGB information. The resulting aligned RGB
features are finally fused with the hyperspectral representations to reconstruct
the high-resolution HSI.

\noindent\textbf{Global Alignment.}
We first compensate for the dominant displacement between the two camera views by estimating a dense optical flow from the hyperspectral view to the RGB view. Unlike a global homography, a dense flow can represent the depth-dependent parallax produced by the two-camera geometry. Since optical-flow estimators operate on images with comparable appearance, the low-resolution HSI is spatially upsampled and converted to a pseudo-RGB image using three representative spectral bands,
\begin{equation}
    \mathbf{Y}^{rgb}
    =
    P\!\left(
        \mathrm{Up}(\mathbf{Y})
    \right),
\end{equation}
where $\mathrm{Up}(\cdot)$ denotes bicubic upsampling to the spatial resolution of $\mathbf{R}$ and $P(\cdot)$ denotes the spectral band selection operator. The pseudo-RGB image and the RGB guide are provided to RAFT~\cite{teed2020raft}, which estimates the dense displacement field
\begin{equation}
    \boldsymbol{\Phi}
    =
    \mathrm{RAFT}
    \left(
        \mathbf{Y}^{rgb},
        \mathbf{R}
    \right).
\end{equation}
The RGB observation is then backward warped to obtain the globally aligned
guide
\begin{equation}
    \mathbf{R}^{g}
    =
    \mathcal{W}
    \left(
        \mathbf{R},
        \boldsymbol{\Phi}
    \right).
\end{equation}
This stage corrects the dominant inter-camera displacement, while finer
geometric discrepancies are handled in feature space.

\noindent\textbf{Multiscale Embedding.}
The hyperspectral observation $\mathbf{Y}$ and the globally aligned RGB guide
$\mathbf{R}^{g}$ are projected at three spatial resolutions. The resulting
features are denoted by $\mathbf{Y}_n$ and $\mathbf{R}_n$, with
$n\in{1,2,3}$ corresponding to full, half, and quarter resolution,
respectively. Multiscale processing has proven effective for RGB-guided HSI
super-resolution by enabling spatial and spectral information to interact across
different resolutions~\cite{kolyszko2026multiscale}. In TANGO, this hierarchy
also supports coarse-to-fine correspondence refinement: coarser representations
provide a larger effective receptive field for handling residual geometric
discrepancies, while finer scales allow more precise spatial matching. The HSI representation is obtained through bilinear upsampling followed by a $3\times3$ convolution, whereas the RGB stream is progressively downsampled and projected to the corresponding feature dimensionality by a $3\times3$ convolution. The resulting
representations are then provided to the local alignment stage.

\noindent\textbf{Local Alignment with Learned Gating.}
Although the global optical-flow stage compensates for the dominant displacement
between the two camera views, it cannot fully resolve depth-dependent parallax,
local perspective variations, and occlusions. These effects are spatially
varying and are therefore more effectively handled after feature extraction.
TANGO consequently performs a second alignment directly in feature space,
progressively refining residual correspondences from the coarsest to the finest
resolution. At each scale, local alignment is formulated as a local cross-attention between the two modalities: the HSI features act as queries, while the RGB features provide keys and values. For a spatial position $p$, local correspondence weights are
computed within a neighborhood $\Omega_r$ as
\begin{equation}
    a_{p,o}
    =
    \operatorname*{softmax}_{o\in\Omega_r}
    \left(
        \tau\,
        \mathbf{q}_{p}^{\top}
        \mathbf{k}_{p+o}
        +
        \beta\,\mathbb{I}[o=0]
    \right),
    \label{eq:local_attention}
\end{equation}
where $\mathbf{q}$ and $\mathbf{k}$ are learned projections of
$\mathbf{Y}_n$ and $\mathbf{R}_n$, respectively, $\tau$ is a learnable
temperature parameter, and $\beta$ introduces a prior toward zero residual displacement.
This prior is consistent with the role of the global alignment stage, which is
expected to remove most of the inter-camera motion before local refinement.

The locally aligned RGB representation is obtained by aggregating candidate
features according to the learned correspondence distribution,
\begin{equation}
    \tilde{\mathbf{R}}_n(p)
    =
    \sum_{o\in\Omega_r}
    a_{p,o}\mathbf{R}_n(p+o),
\end{equation}
while the expected residual displacement is obtained from the same distribution,
\begin{equation}
    \Delta\mathbf{u}_n(p)
    =
    \sum_{o\in\Omega_r}
    a_{p,o}\,o.
\end{equation}
This soft correspondence formulation avoids committing to a single hard match
and is therefore better suited to ambiguous local structures and imperfect
global registration.

Not all local correspondences are equally reliable, particularly in regions
affected by occlusions or geometric ambiguity. To account for this uncertainty,
TANGO exploits the local correspondence response itself as a confidence cue,
following the common interpretation of matching confidence as an indicator of
correspondence reliability~\cite{kim2019laf, luo2020attention}. Following
this principle, TANGO predicts a gate from the HSI feature, the locally aligned RGB representation, and the maximum local attention response,
\begin{equation}
    g_n
    =
    \sigma
    \left(
        \phi_n
        \left[
            \mathbf{Y}_n,
            \tilde{\mathbf{R}}_n,
            m_n
        \right]
    \right),
    \qquad
    m_n(p)=\max_o a_{p,o}.
\end{equation}
where $\phi_n$ consists of two $3\times3$ convolutions separated by a LeakyReLU, and $\sigma$ is the sigmoid function.
The maximum attention response acts as an implicit measure of local matching
confidence: a dominant response indicates the presence of a well-defined
correspondence, whereas a low maximum response suggests a more ambiguous local
match. The gate therefore learns to modulate the contribution of the aligned
RGB features according to both cross-modal consistency and correspondence
reliability. The feature passed to the reconstruction backbone is
\begin{equation}
    \hat{\mathbf{R}}_n
    =
    g_n\odot\tilde{\mathbf{R}}_n.
\end{equation}

Alignment proceeds from $n=3$ to $n=1$. At the coarsest scale, a $9\times9$
window provides a wider search range for larger residual displacements. The
estimated displacement is then upsampled and used to warp the RGB features at
the following scale, where a $5\times5$ neighborhood estimates a finer residual
correction. This coarse-to-fine strategy progressively narrows the
correspondence search as the spatial resolution increases, while the gating
mechanism independently controls the reliability of the transferred RGB
information at each scale.

\noindent\textbf{Multiscale Spatial-Spectral Fusion.}
The aligned features $\hat{\mathbf{R}}_n$ are combined with the corresponding
hyperspectral features $\mathbf{Y}_n$ using the multiscale fusion backbone of
LRTN~\cite{liu2025low}. The encoder contains complementary spatial and spectral
streams: Spatial Transformers transfer high-frequency spatial information from
the aligned RGB representations, while Spectral Transformers preserve and
propagate the spectral information contained in the HSI.

At the coarsest level, the two streams are merged and processed by a
Prior-Guided Transformer. The decoder then progressively restores the spatial
resolution using PixelShuffle and skip connections from the corresponding
encoder stages. Prior-Guided Transformers further combine the aligned RGB and
HSI representations at each scale. Unlike the original LRTN formulation,
every RGB representation used by the fusion backbone is replaced by the
aligned and gated feature $\hat{\mathbf{R}}_n$, ensuring that RGB spatial
information is injected only after correspondence refinement.

The final feature representation is projected to $C$ spectral channels and
combined with the upsampled HSI through a residual connection,
\begin{equation}
    \hat{\mathbf{X}}
    =
    \mathrm{Conv}_{1\times1}
    \left[
        \mathbf{F}_1,
        \mathbf{Y}_1,
        \hat{\mathbf{R}}_1
    \right]
    +
    \mathbf{Y}_1.
\end{equation}

\noindent\textbf{Training Objective.} TANGO is trained end to end with an
$\ell_1$ reconstruction loss and an auxiliary alignment loss on the globally
warped RGB, following SSC-HSR~\cite{zhang2025unaligned},
\begin{equation}
    \mathcal{L}
    =
    \big\|\hat{\mathbf{X}}-\mathbf{X}\big\|_1
    +
    \lambda_{\mathrm{align}}
    \big\|\mathbf{R}^{g}-\mathbf{X}^{rgb}\big\|_1 ,
\end{equation}
where $\mathbf{X}$ is the high-resolution hyperspectral reference and
$\mathbf{X}^{rgb}$ is its pseudo-RGB, built from the same bands as
$\mathbf{Y}^{rgb}$ and histogram-matched to $\mathbf{R}$. We set
$\lambda_{\mathrm{align}}=1$. RAFT is fine-tuned jointly with the
reconstruction network, so the global correspondence estimation adapts to the
RGB--HSI task.
\section{Experiments}
\label{sec:experiments}

\subsection{Experimental Settings}
\noindent\textbf{Datasets and Metrics.}
We evaluate TANGO on the proposed DUET dataset and on
REAL~\cite{lai2024hyperspectral}.
DUET contains 403 real RGB--HSI pairs acquired with separate cameras,
whereas REAL contains 57 unregistered hyperspectral pairs, with RGB
guidance synthesized from one hyperspectral observation. DUET results are reported at $\times4$, $\times8$, and $\times16$; for REAL, we report $\times8$ in the main paper and $\times4$ and $\times16$ in the supplementary material.
Reconstruction quality is evaluated using Peak Signal-to-Noise Ratio (PSNR),
Structural Similarity Index (SSIM)~\cite{Wang2004SSIM}, and Spectral Angle
Mapper (SAM)~\cite{Yuhas1992SAM}. Higher PSNR and SSIM and lower SAM indicate
better reconstruction quality. 

\noindent\textbf{Implementation Details.}
TANGO is implemented in PyTorch and trained on a single NVIDIA TITAN Xp GPU for
300 epochs with a batch of four $128\times128$ high-resolution crops. The HR-HSI, LR-HSI, and RGB crops are extracted at the same random location, so that cropping does not add misalignment beyond the one introduced by the acquisition. RAFT, initialized from the RAFT-Things weights~\cite{teed2020raft}, estimates the global flow on the whole crop, where enough spatial context is available to capture large inter-camera displacements. The warped RGB and the HSI are then divided into non-overlapping $64\times64$ tiles, which the rest of the network processes independently, following~\cite{liu2025low}. At inference, RAFT operates on the full image, while the local
alignment and fusion stages use $64\times64$ tiles with a stride of 32, and
predictions in overlapping regions are averaged. The network is optimized with
Adam (learning rate $4\times10^{-4}$, weight decay $10^{-8}$) and cosine
annealing, while RAFT is fine-tuned with a reduced learning rate of $10^{-5}$.

\noindent\textbf{Compared Methods.}
We compare TANGO with Bicubic interpolation, the single-image HSI
super-resolution methods SSPSR~\cite{FPB2020}, Bi-3DQRNN~\cite{9351612},
and ESSAformer~\cite{EssaFormer}, and the RGB-guided methods
LRTN~\cite{liu2025low}, HSIFN~\cite{lai2024hyperspectral},
M2DTN~\cite{Dong_Xu_Qu_Hou_2024}, SSC-HSR~\cite{zhang2025unaligned}, and
UAFL~\cite{zhang2026enhancing}.
For learning-based baselines, we consider both the original training
configuration and the same training protocol adopted for TANGO, retaining the
best validation result.

\subsection{Main Results}
\noindent\textbf{Results on DUET.}
Table~\ref{tab:DUET_results} shows that TANGO achieves the best performance across all metrics and scale factors. Compared with the strongest competitor, SSC-HSR, TANGO improves PSNR by 2.235 dB, 1.713 dB, and 2.285 dB at $\times4$, $\times8$, and $\times16$, respectively. The gains are consistent for SSIM and SAM, indicating improved spatial reconstruction and spectral fidelity. At $\times16$, TANGO reaches 32.418 dB PSNR and 0.9047 SSIM while reducing SAM to 0.05725, highlighting the benefit of progressive RGB--HSI alignment under severe spatial degradation. These results are obtained with 8.887\,M parameters and 0.80\,s per $512\times512$ image, about 2.5$\times$ fewer FLOPs and a lower inference time than SSC-HSR (0.92\,s). Lighter fusion models such as M2DTN, UAFL, and LRTN are faster, but achieve at least 4.6\,dB lower PSNR at $\times4$. An analysis of the trade-off between accuracy and computational cost, including the Pareto front with respect to FLOPs and PSNR, is reported in 
the supplementary material. Figure~\ref{fig:DUET_qualitative} further confirms the quantitative results. TANGO produces cleaner error maps around object boundaries, fine contours, and text, while limiting artifacts in regions affected by cross-camera misalignment. The corresponding reconstructions are shown in 
the supplementary material.

\begin{table*}[t]
\centering
\caption{Quantitative comparison on DUET at different scale factors.
PSNR is reported in dB. Model complexity and inference time are reported for
the reference $\times4$ configuration on a $512\times512$ image (batch size 1,
NVIDIA TITAN Xp; tile-based methods process tiles in batches).
$\uparrow$ indicates that higher values are better, while $\downarrow$
indicates that lower values are better. The best and second-best results are
highlighted in \textbf{bold} and \underline{underline}, respectively.}
\label{tab:DUET_results}

\small
\setlength{\tabcolsep}{4pt}

\begin{tabular}{l ccc ccc ccc lll}
\toprule

\multirow{2}{*}{\textbf{Method}} &
\multicolumn{3}{c}{\textbf{Scale Factor} $\times4$} &
\multicolumn{3}{c}{\textbf{Scale Factor} $\times8$} &
\multicolumn{3}{c}{\textbf{Scale Factor} $\times16$} &
\multirow{2}{*}{\shortstack{\textbf{Params}\\\textbf{[M]}}} &
\multirow{2}{*}{\shortstack{\textbf{FLOPs}\\\textbf{[G]}}} &
\multirow{2}{*}{\shortstack{\textbf{Time}\\\textbf{[s]}}} \\

\cmidrule(lr){2-4}
\cmidrule(lr){5-7}
\cmidrule(lr){8-10}

&
PSNR$\uparrow$ & SSIM$\uparrow$ & SAM$\downarrow$ &
PSNR$\uparrow$ & SSIM$\uparrow$ & SAM$\downarrow$ &
PSNR$\uparrow$ & SSIM$\uparrow$ & SAM$\downarrow$ &
& & \\

\midrule

Bicubic
& 31.140 & 0.8599 & 0.04541
& 26.603 & 0.7462 & 0.07357
& 23.140 & 0.6616 & 0.11375
& \multicolumn{1}{c}{--} & \multicolumn{1}{c}{--} & 0.01 \\

SSPSR~\cite{FPB2020}
& 38.475 & 0.9515 & \underline{0.02812}
& 31.274 & 0.8512 & \underline{0.04908}
& 25.879 & 0.7375 & 0.07890
& 26.08 & 21066.1 & 1.72 \\

Bi-3DQRNN~\cite{9351612}
& 37.662 & 0.9441 & 0.03324
& 30.712 & 0.8369 & 0.05477
& 25.882 & 0.7247 & 0.08839
& \underline{1.288} & 3749.4 & 1.11 \\

ESSAformer~\cite{EssaFormer}
& 37.936 & 0.9478 & 0.03115
& 30.752 & 0.8406 & 0.05381
& 25.729 & 0.7327 & 0.08317
& 11.194 & 23015.6 & 2.14 \\

\midrule

HSIFN~\cite{lai2024hyperspectral}
& 35.931 & 0.9264 & 0.04682
& 32.103 & 0.8753 & 0.06406
& 28.585 & 0.8139 & 0.09318
& 69.922 & 10707.3 & 2.54 \\

M2DTN~\cite{Dong_Xu_Qu_Hou_2024}
& 36.413 & 0.9387 & 0.05545
& 30.613 & 0.8387 & 0.06644
& 27.318 & 0.7822 & 0.08671
& 7.823 & 1001.0 & \textbf{0.24} \\

LRTN~\cite{liu2025low}
& 39.046 & 0.9597 & 0.03127
& 33.292 & 0.9004 & 0.05001
& 29.058 & 0.8329 & \underline{0.07061}
& 3.551 & \underline{940.6} & 0.45 \\

SSC-HSR~\cite{zhang2025unaligned}
& \underline{41.437} & \underline{0.9757} & 0.03505
& \underline{36.403} & \underline{0.9429} & 0.05745
& \underline{30.133} & \underline{0.8634} & 0.08276
& 27.451 & 3475.9 & 0.92 \\

UAFL~\cite{zhang2026enhancing}
& 36.080 & 0.9275 & 0.04276
& 30.861 & 0.8411 & 0.06559
& 26.566 & 0.7439 & 0.10137
& \textbf{0.358} & \textbf{143.1} & \underline{0.41} \\

\textbf{TANGO}
& \textbf{43.672} & \textbf{0.9833} & \textbf{0.02467}
& \textbf{38.116} & \textbf{0.9560} & \textbf{0.03762}
& \textbf{32.418} & \textbf{0.9047} & \textbf{0.05725}
& 8.887 & 1398.5 & 0.80 \\

\bottomrule
\end{tabular}
\end{table*}

\begin{figure*}[t]
    \centering
    \includegraphics[width=\linewidth]{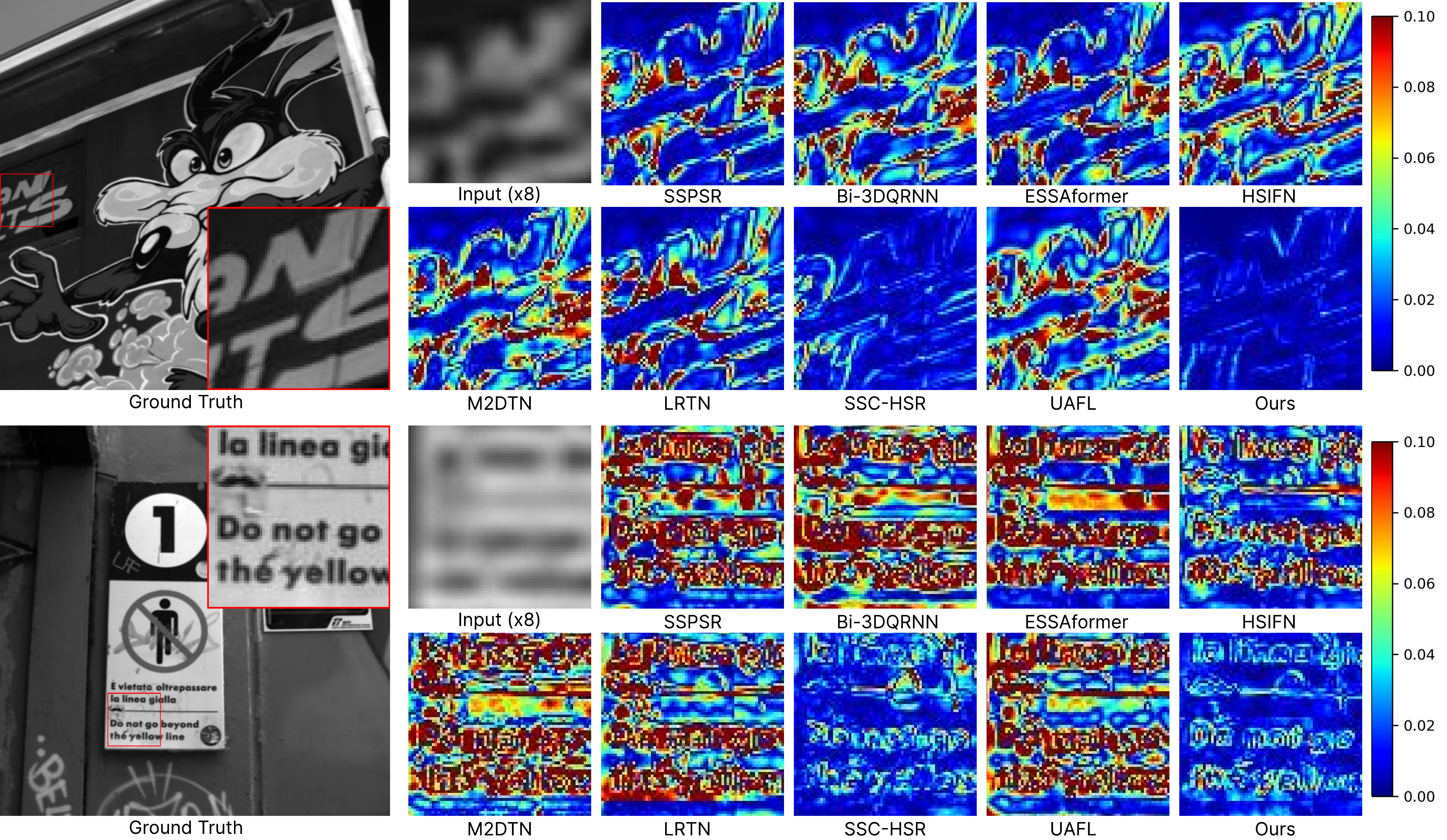}
    \caption{Qualitative comparison on DUET at scale factor $\times8$.
    The first column shows the ground-truth HSI and the selected regions of interest. The low-resolution input and the absolute reconstruction error maps of the compared methods are shown on the right. Lower values indicate more accurate reconstruction.}
    \label{fig:DUET_qualitative}
\end{figure*}

\begin{figure*}[t]
    \centering
    \includegraphics[width=\linewidth]{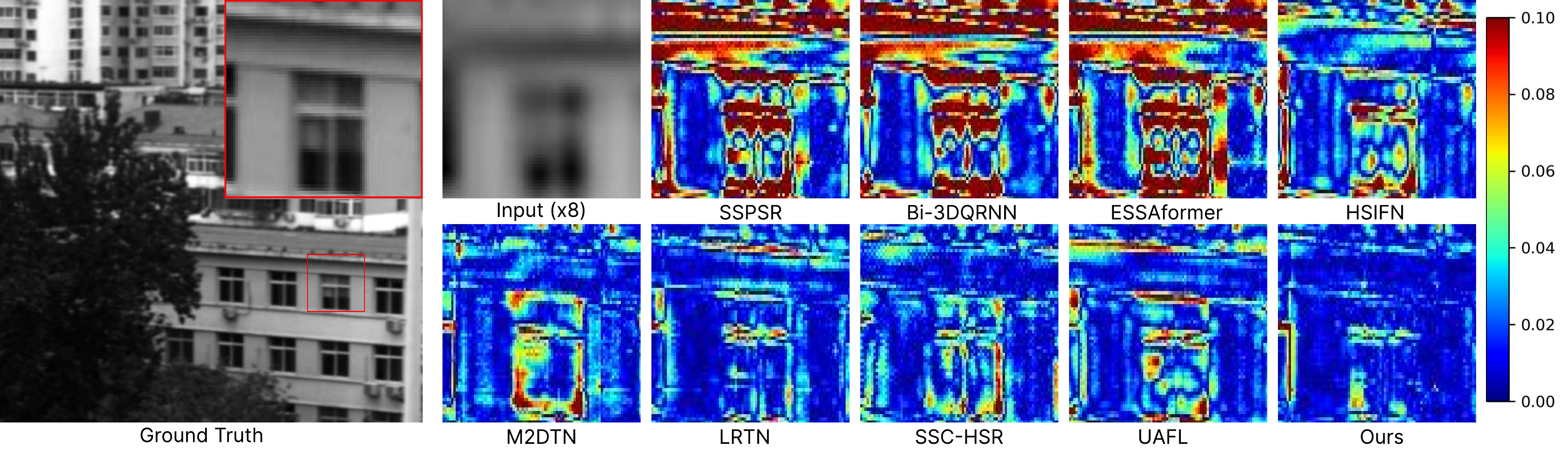}
    \caption{Qualitative comparison on the REAL dataset at $\times8$.
    The first column shows the ground-truth HSI and the selected regions of interest. The low-resolution input and the absolute reconstruction error maps of the compared methods are shown on the right. Lower values indicate more accurate reconstruction.}
    \label{fig:real_qualitative}
\end{figure*}

\noindent\textbf{Results on REAL.}
Table~\ref{tab:real_results} shows that TANGO achieves the best performance across all metrics on REAL at $\times8$, reaching 37.368 dB PSNR, 0.9410 SSIM, and 0.04537 SAM. Compared with the strongest competitor, LRTN, TANGO improves PSNR by 0.944 dB while also improving SSIM and SAM, demonstrating that the proposed alignment strategy generalizes to a different unregistered acquisition setting. Figure~\ref{fig:real_qualitative} confirms this trend, with TANGO producing cleaner error maps around structural edges and fine architectural details, indicating more reliable use of RGB guidance under cross-view misalignment. All methods are retrained on REAL under the same protocol, and results at $\times4$ and $\times16$ are provided in 
the supplementary material.

\begin{table}[t]
\centering
\caption{Quantitative comparison on the REAL dataset at $\times8$.
PSNR is reported in dB. $\uparrow$ indicates that higher values are better,
while $\downarrow$ indicates that lower values are better. The best and
second-best results are highlighted in \textbf{bold} and
\underline{underline}, respectively.}
\label{tab:real_results}

\small
\setlength{\tabcolsep}{6pt}

\begin{tabular}{l c c c}
\toprule
\textbf{Method} &
\textbf{PSNR$\uparrow$} &
\textbf{SSIM$\uparrow$} &
\textbf{SAM$\downarrow$} \\
\midrule

Bicubic
& 28.675 & 0.7810 & 0.05743 \\

SSPSR~\cite{FPB2020}
& 32.404 & 0.8496 & 0.04912 \\

Bi-3DQRNN~\cite{9351612}
& 31.926 & 0.8379 & 0.05433 \\

ESSAformer~\cite{EssaFormer}
& 31.443 & 0.8304 & 0.06384 \\

\midrule

HSIFN~\cite{lai2024hyperspectral}
& 33.181 & 0.8795 & 0.07013 \\

M2DTN~\cite{Dong_Xu_Qu_Hou_2024}
& 34.027 & 0.9121 & 0.06966 \\

LRTN~\cite{liu2025low}
& \underline{36.424} & \underline{0.9236} & \underline{0.04909} \\

SSC-HSR~\cite{zhang2025unaligned}
& 35.532 & 0.9221 & 0.05945 \\

UAFL~\cite{zhang2026enhancing}
& 34.564 & 0.9038 & 0.05593 \\

\textbf{TANGO}
& \textbf{37.368} & \textbf{0.9410} & \textbf{0.04537} \\

\bottomrule
\end{tabular}
\end{table}

\subsection{Ablation Study}
Table~\ref{tab:tango_ablation} analyzes the contribution of the global and local alignment stages on DUET at $\times8$, while the complete ablation at $\times4$, $\times8$, and $\times16$ is provided in 
the supplementary material. Without global alignment, local alignment alone improves PSNR by 0.35\,dB with only 0.079\,M additional parameters, showing that feature matching can partially compensate for the cross-camera misalignment. A frozen pretrained RAFT yields a much larger gain of 2.05\,dB, confirming that compensating for the dominant displacement before fusion substantially simplifies the reconstruction, and local alignment still adds 0.34\,dB on top of it. Fine-tuning RAFT end to end gives a further 2.55\,dB, indicating that a generic optical-flow estimator is not fully adapted to RGB--HSI correspondence. On top of this configuration, local alignment provides an additional 0.228\,dB. Local alignment consistently reduces SAM in all three configurations. Global and local alignment therefore play complementary roles: the former corrects the dominant inter-camera displacement, whereas the latter refines residual local inconsistencies. Table~\ref{tab:gate_ablation} isolates the role of the gate by fixing $g_n = 1$ in the full model and keeping the same training recipe. Without the gate, local alignment no longer improves over the global-only configuration, reaching 37.867\,dB against 37.888\,dB, which indicates that injecting the attended RGB features without modulation is harmful where correspondences are unreliable. With only 0.063\,M parameters, the gate restores the benefit of local alignment, and the complete TANGO reduces the reconstruction error most on pixels without a reliable correspondence, by 6.2\% with respect to the global-only configuration. Results at all scale factors and a per-region analysis of the gate are provided in 
the supplementary material.

\begin{table}[t]
\centering
\caption{Ablation study on DUET at scale factor $\times8$.
$\times$ denotes disabled global alignment, \faSnowflake~denotes frozen
pretrained RAFT, and \checkmark ~denotes end-to-end optimized global alignment.}
\label{tab:tango_ablation}

\small
\setlength{\tabcolsep}{4.2pt}

\begin{tabular}{c c c c c c}
\toprule
\textbf{Global} &
\textbf{Local} &
\textbf{PSNR$\uparrow$} &
\textbf{SSIM$\uparrow$} &
\textbf{SAM$\downarrow$} &
\textbf{Params [M]} \\
\midrule

\multirow{2}{*}{$\times$}
& $\times$
& 33.292 & 0.9004 & 0.05001 & 3.551 \\

&
\checkmark
& 33.644 & 0.9059 & 0.04914 & 3.630 \\

\midrule

\multirow{2}{*}{\faSnowflake}
& $\times$
& 35.339 & 0.9330 & 0.04221 & 3.551 \\

&
\checkmark
& 35.676 & 0.9367 & 0.04060 & 3.630 \\

\midrule

\multirow{2}{*}{\checkmark}
& $\times$
& 37.888 & 0.9534 & 0.03942 & 8.809 \\

&
\checkmark
& \textbf{38.116} & \textbf{0.9560} & \textbf{0.03762} & 8.887 \\

\bottomrule
\end{tabular}
\end{table}

\begin{table}[t]
\centering
\caption{Effect of the gate in the local alignment on DUET at scale factor $\times8$. Both variants use end-to-end optimized global alignment.}
\label{tab:gate_ablation}

\small
\setlength{\tabcolsep}{4.2pt}

\begin{tabular}{l c c c c}
\toprule
\textbf{Local alignment} &
\textbf{PSNR$\uparrow$} &
\textbf{SSIM$\uparrow$} &
\textbf{SAM$\downarrow$} &
\textbf{Params [M]} \\
\midrule
w/o gate & 37.867 & 0.9528 & 0.03886 & 8.824 \\
w/ gate  & \textbf{38.116} & \textbf{0.9560} & \textbf{0.03762} & 8.887 \\
\bottomrule
\end{tabular}
\end{table}
\section{Conclusion}
\label{sec:conclusion}

We presented DUET, a large-scale dataset of real two-camera RGB--HSI acquisitions, and TANGO, a method for unregistered RGB-guided hyperspectral super-resolution under these realistic conditions. DUET provides real RGB--HSI pairs independently captured with separate RGB and hyperspectral cameras, preserving the geometric and cross-sensor discrepancies that arise in practical systems. Building on this setting, TANGO combines global alignment, coarse-to-fine local feature refinement, and a learned gating mechanism to exploit RGB guidance while reducing the effect of unreliable cross-modal correspondences. Experimental results on both DUET and REAL datasets show that TANGO achieves state-of-the-art reconstruction performance across all evaluated scale factors. The ablation study further confirms that global and local alignment play complementary roles, and that jointly adapting the global correspondence estimator to the reconstruction task is important for robust performance. Overall, the proposed dataset and framework provide a step toward evaluating and developing RGB-guided hyperspectral super-resolution methods under more realistic acquisition conditions.

{
    \small
    \bibliographystyle{ieeenat_fullname}
    \bibliography{main}
}

\clearpage
\setcounter{page}{1}
\maketitlesupplementary
 
This supplementary material complements the main paper with further details and results. \Cref{sec:supp_dataset} describes the DUET acquisition setup and preprocessing, and \cref{sec:supp_misalignment} analyzes the cross-camera misalignment. \Cref{sec:supp_results} reports additional qualitative results on DUET, the trade-off between accuracy and complexity, and the complete results on REAL, while \cref{sec:supp_ablation} presents the complete ablation study, including an analysis of the gate.
 
\section{DUET Dataset}
\label{sec:supp_dataset}
 
\subsection{Acquisition Setup}
\label{sec:supp_setup}

\begin{figure}[t]
    \centering
    \includegraphics[width=\linewidth]{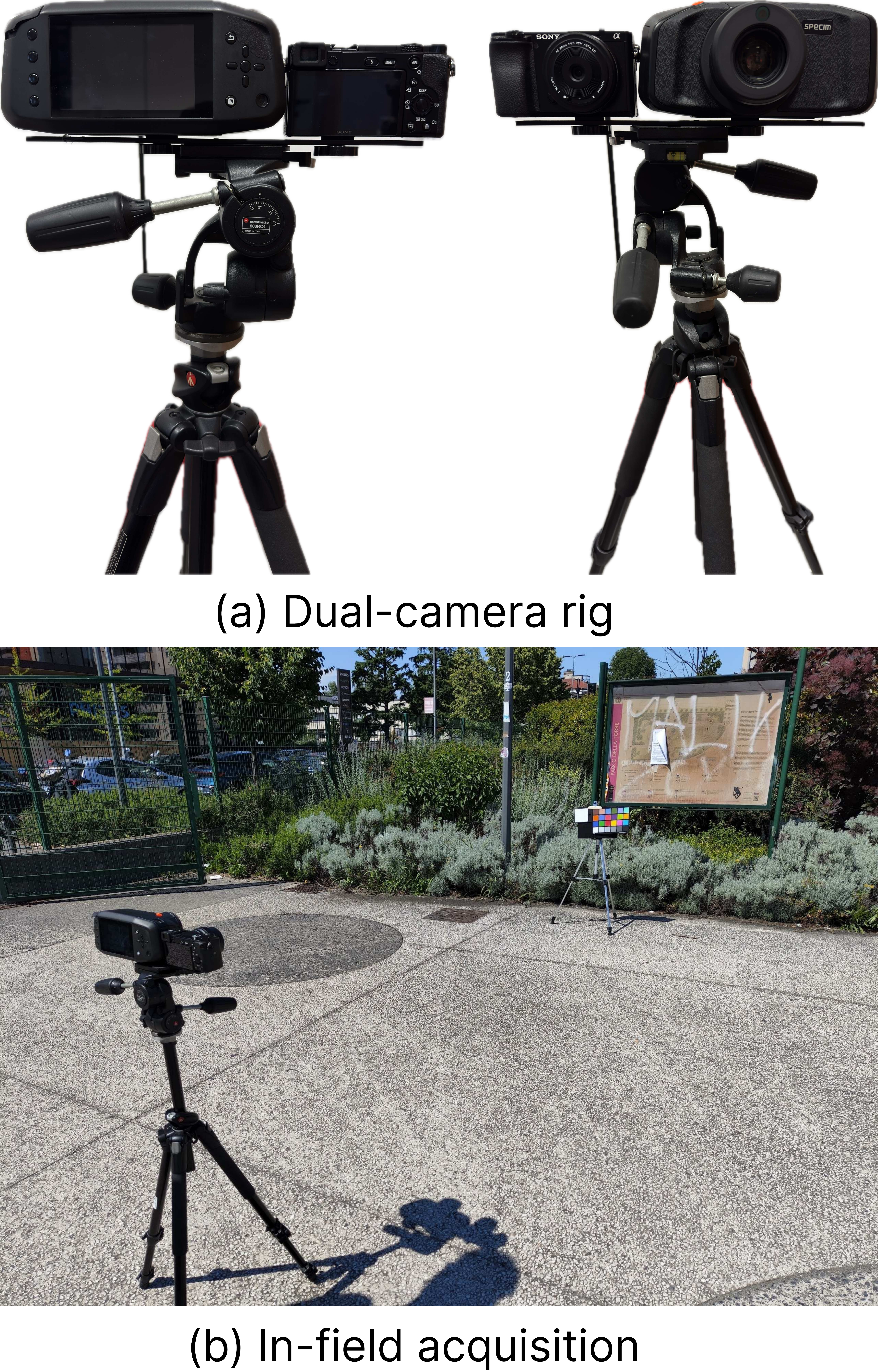}
    \caption{DUET acquisition setup. (a) Dual-camera rig, shown from the rear and from the front, with approximately 15\,cm between the optical centers. (b) The rig during an outdoor acquisition; the calibration target is shown for illustration only and is not present in the dataset scenes.}
    \label{fig:supp_setup}
\end{figure}

\Cref{fig:supp_setup} shows the acquisition setup. The Specim IQ hyperspectral camera and the Sony $\alpha$6100 RGB camera are mounted side by side on a common plate on a tripod, with the Specim IQ on the left and the Sony on the right when viewed from behind. This configuration results in a horizontal baseline of approximately 15\,cm between the optical centers, which is kept fixed for all scenes. The two cameras are operated independently: the Sony acquires RAW images at approximately $6000\times4000$ pixels, while the Specim IQ acquires $512\times512$ cubes with 204 bands between 397 and 1004\,nm over a field of view of 31$^\circ$. Since the Specim IQ is a scanning camera, each cube is acquired over a finite exposure time, during which the scene and the illumination must remain stable; scenes with moving objects were therefore avoided. Before each scene, a white reference panel is acquired with the Specim IQ and used to convert the hyperspectral measurements to reflectance; the panel is removed before the acquisition and does not appear in the dataset images.

Since the two cameras observe the scene from different positions, their images are related by a depth-dependent displacement rather than by a global transformation. Under a parallel-axis approximation, the Specim IQ field of view of 31$^\circ$ over 512\,px corresponds to a focal length of $f = 256/\tan(15.5^\circ) \approx 923$\,px, and the disparity of a point at distance $Z$ is $d = fB/Z \approx 138/Z$\,px, with $Z$ in meters. \Cref{tab:supp_disparity} reports the resulting disparity at representative distances. Since the field-of-view matching of the main paper (\cref{sec:DUET_dataset}) removes a single scale and translation, what remains is mainly the \emph{difference} in disparity between objects at different depths: for example, a foreground object at 3\,m in front of a background at 30\,m is displaced by about 41\,px relative to it. Typical DUET scenes span depths from about one to a few tens of meters, so this relative displacement ranges from a few pixels to several tens of pixels, consistently with the misalignment statistics of \cref{sec:supp_misalignment}. In addition, the horizontal baseline causes occlusions along depth discontinuities: since the hyperspectral camera is on the left, the background immediately to the left of a foreground object is visible in the hyperspectral image but hidden in the RGB view, which produces the occlusion strips shown in \cref{fig:supp_corr_maps}.

\begin{table}[ht]
\centering
\caption{Approximate disparity between the two views as a function of the scene distance $Z$, under a parallel-axis approximation, in pixels of the $512\times512$ hyperspectral image.}
\label{tab:supp_disparity}
\small
\begin{tabular}{l ccccc}
\toprule
$Z$ [m] & 1 & 3 & 10 & 30 & 100 \\
\midrule
Disparity [px] & 138 & 46 & 14 & 5 & 1.4 \\
\bottomrule
\end{tabular}
\end{table}
 
\subsection{Additional Examples}
\label{sec:supp_examples}

\begin{figure*}[t]
    \centering
    \includegraphics[width=\linewidth]{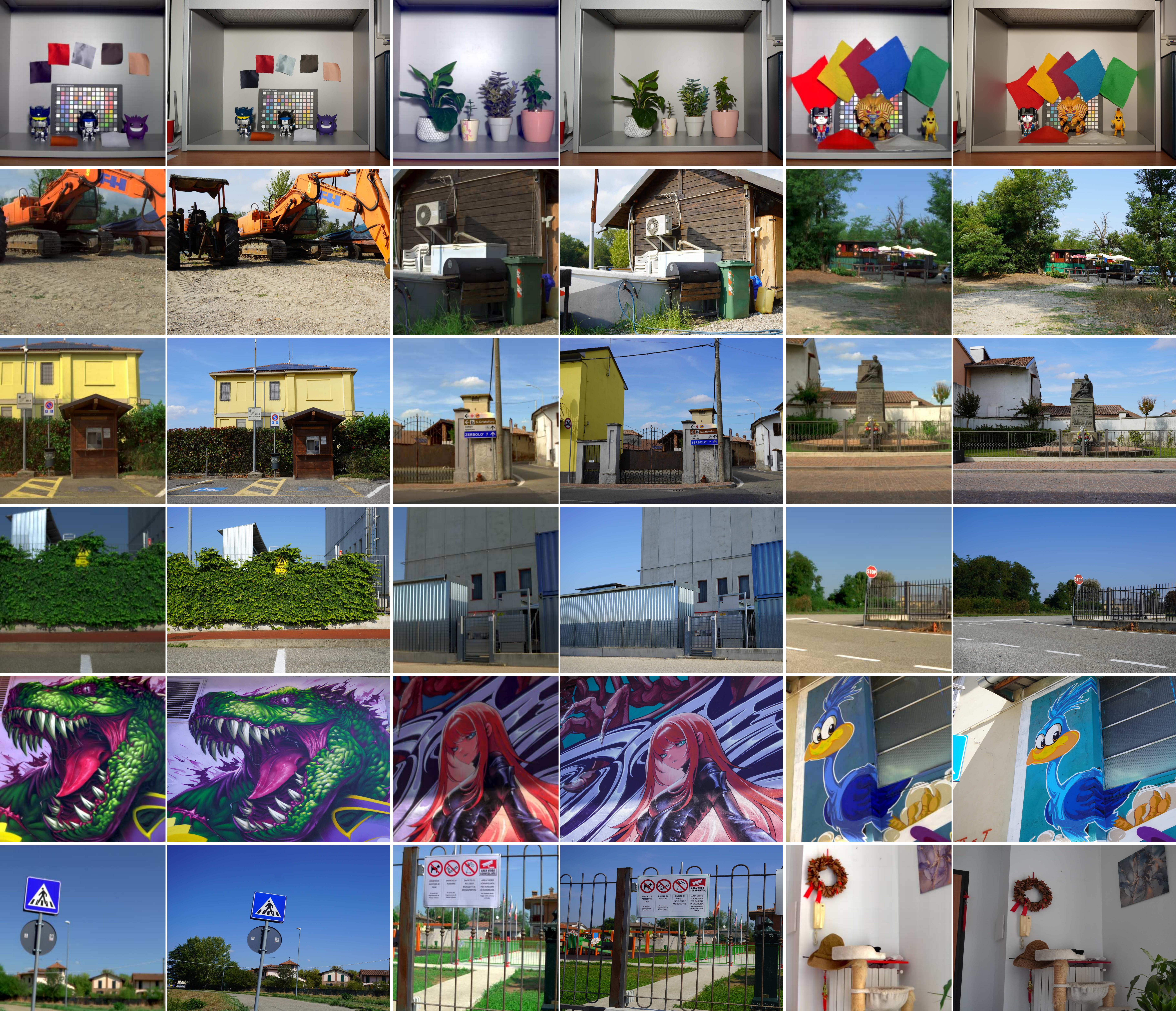}
    \caption{Additional RGB--HSI pairs from DUET. RGB images are the original in-camera JPEGs with the full Sony field of view, and hyperspectral cubes are rendered as pseudo-RGB images from the bands at 650, 560, and 470\,nm, contrast-stretched for visualization only.}
    \label{fig:supp_examples}
\end{figure*}

\Cref{fig:supp_examples} shows additional RGB--HSI pairs from DUET, covering both indoor and outdoor scenes. Indoor scenes are recorded under controlled tungsten illumination and include objects with a wide range of colors and materials, such as fabrics, figurines, color charts, and plants. Outdoor scenes are captured under natural daylight and include buildings, vegetation, vehicles, street art, and road signs, at distances ranging from a few meters to several tens of meters. As in \cref{fig:DUET_samples}, the RGB images cover a wider field of view than the hyperspectral images and are acquired from a different viewpoint, so that each pair exhibits the cross-camera discrepancies addressed in the main paper.

\subsection{Field-of-View Matching}
\label{sec:supp_preprocessing}

\begin{figure}[t]
    \centering
    \includegraphics[width=\columnwidth]{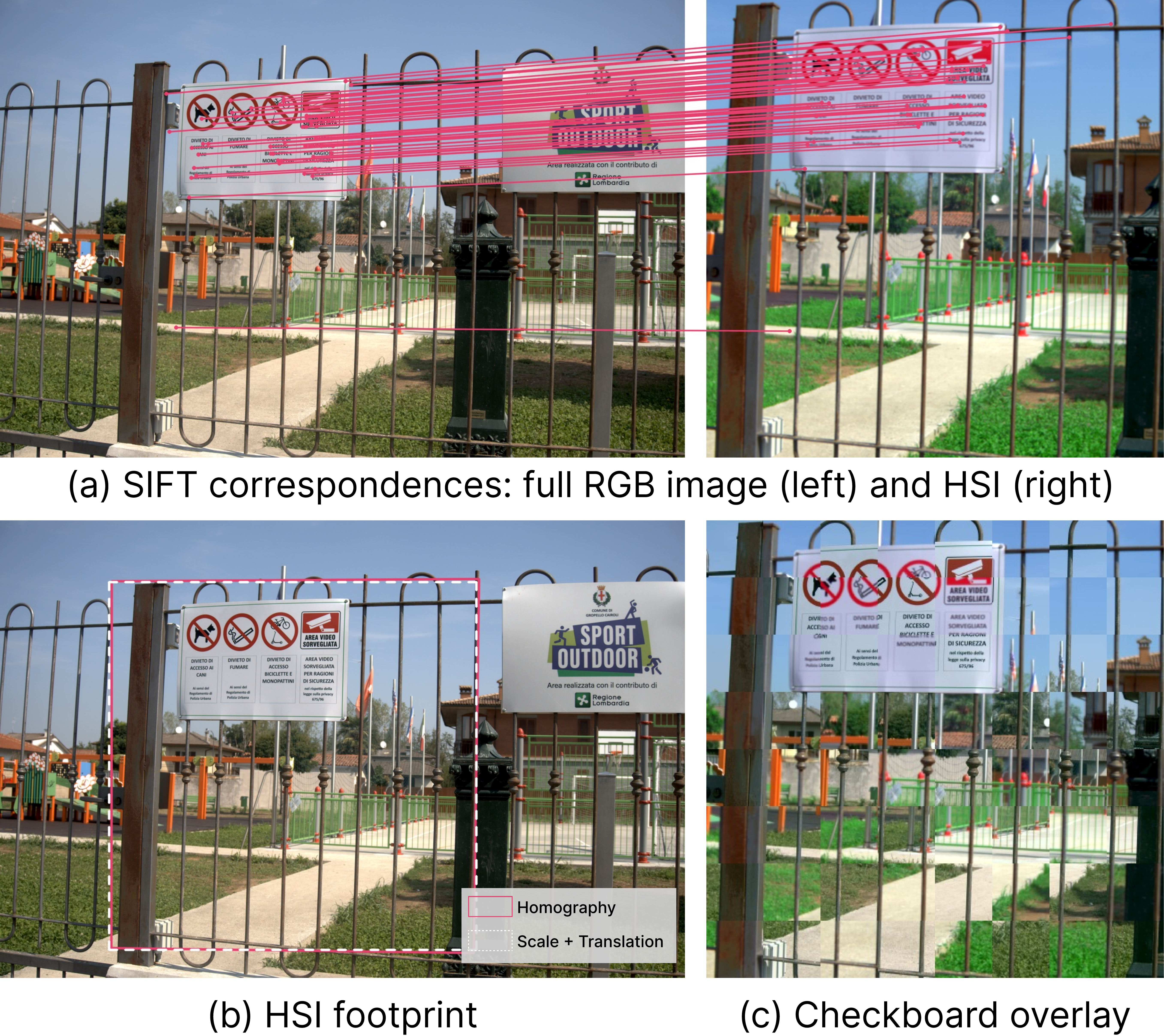}
    \caption{Field-of-view matching on a DUET scene. (a) SIFT correspondences retained by RANSAC between the full RGB image and the HSI. (b) HSI footprint projected through the estimated homography and fitted square region. (c) Checkerboard overlay of the cropped RGB image and the HSI, showing the residual misalignment preserved by the procedure. For visualization only, the linear RGB image is white-balanced and both modalities are contrast-stretched.}
    \label{fig:supp_preprocessing}
\end{figure}

The Sony camera has a wider field of view and a much higher resolution than the Specim IQ, so each RGB image covers a larger area of the scene than the corresponding hyperspectral cube. Before constructing the benchmark, the two views are therefore brought to a common field of view with the procedure described in \cref{sec:DUET_dataset}, which is illustrated in \cref{fig:supp_preprocessing} on a representative scene.

SIFT features are extracted from the full Sony image and from the hyperspectral pseudo-RGB, matched with FLANN, and filtered with Lowe's ratio test. The retained correspondences (a) are used within RANSAC to estimate a homography from the hyperspectral to the RGB view. Rather than warping the RGB image with this homography, which would partially register the two views and introduce resampling artifacts, the homography is approximated by an isotropic scale and translation, fitted with least squares to a regular grid of hyperspectral pixels mapped through it. As shown in (b), the resulting square region closely matches the projected hyperspectral footprint, so the approximation removes the field-of-view difference without compensating for the remaining geometric discrepancies. The square region is then cropped from the RGB image and resized to $512\times512$ pixels.

The checkerboard overlay of the cropped RGB image and the hyperspectral image (c) shows that the two views cover the same field of view, but are not registered. Residual rotation, perspective distortion, and depth-dependent parallax remain visible, for example on the doubled sign and on the shifted foreground post, which lie at different depths from the background. These discrepancies are intentionally preserved, since they reflect the misalignment that a real two-camera system produces and that unregistered RGB-guided methods must handle.

\subsection{Degradation Protocol}
\label{sec:supp_degradation}

The $512\times512$ hyperspectral cubes serve as high-resolution references, and the corresponding low-resolution inputs are synthesized from them. Each high-resolution cube is first min--max normalized to $[0,1]$ over all bands. Low-resolution observations are then generated with the protocol of Real-HSI-Fusion~\cite{lai2024hyperspectral}: the cube is convolved band by band with an $8\times8$ Gaussian kernel with standard deviation $\sigma = 3$, using replicated borders, and then decimated by keeping one pixel every $s$ along each spatial dimension. The same kernel is used for all scale factors, yielding low-resolution inputs of $128\times128$, $64\times64$, and $32\times32$ pixels at $\times4$, $\times8$, and $\times16$, respectively, while the 31 spectral bands are left unchanged. Our implementation reproduces the official one used by previous unregistered methods exactly, so that results on REAL and DUET are obtained under the same degradation. The RGB image is not degraded and is provided at the full $512\times512$ resolution as guidance. For methods that require a pre-upsampled input, as well as for the Bicubic baseline, the low-resolution cube is upsampled to $512\times512$ with bicubic interpolation.
 
\section{Misalignment Analysis}
\label{sec:supp_misalignment}

This section quantifies the residual misalignment between the two views of DUET after the field-of-view matching, and shows that it cannot be described by a single global transformation.

\noindent\textbf{Procedure.}
For each of the 403 scenes, we build the pseudo-RGB of the high-resolution hyperspectral cube with the same bands used by TANGO, at approximately 550, 480, and 430\,nm, and match its histogram to the RGB image. We use the high-resolution cube rather than the low-resolution input so that the analysis measures the misalignment of the dataset itself, independently of the scale factor. The optical flow between the two images is estimated in both directions with RAFT~\cite{teed2020raft}, using the public RAFT-Things weights without any fine-tuning on DUET, so that the estimate does not depend on the model under evaluation.

A pixel is marked as having no reliable correspondence when it fails the forward--backward consistency check, \ie, when the forward flow and the backward flow sampled at the corresponding position do not cancel out up to a tolerance proportional to their magnitude. Pixels whose correspondence falls outside the RGB image are also excluded. A homography is then fitted with RANSAC, with a reprojection threshold of 3\,px, to the correspondences given by the flow on a regular grid of the remaining pixels, sampled every 4\,px. The non-planar residual of each pixel is defined as the magnitude of the difference between its flow and the displacement predicted by the homography, and it measures the part of the displacement that no global projective transformation can explain. For each scene, we report the median and 95th percentile of the residual over the pixels with a reliable correspondence, together with the fraction of pixels above 2 and 4\,px and the fraction of pixels without a reliable correspondence.

Since the residual also includes flow estimation errors, it should be interpreted as an estimate of the parallax that cannot be represented by a global transformation rather than as an exact measurement. Similarly, the consistency check may also flag pixels in textureless regions, such as sky or uniform walls, where the flow is ambiguous even without a true occlusion.

\noindent\textbf{Results.}
\Cref{tab:supp_misalignment} summarizes the statistics over the 403 scenes. For the median scene, the 95th percentile of the residual is 11.1\,px, and one scene in ten exceeds 26.4\,px. The 95th percentile is above 4\,px in 72.7\% of the scenes, and in the median scene 22\% of the pixels have a residual larger than 4\,px. For reference, a displacement of 4\,px at full resolution corresponds to one low-resolution pixel at $\times4$ and to half a pixel at $\times8$, which is enough to transfer high-frequency RGB details to wrong positions. These values confirm that a single global transformation cannot register most DUET scenes, and they motivate the design of TANGO: the dense global flow compensates for the dominant displacement, while the $9\times9$ window of the local alignment at quarter resolution, which spans $\pm16$\,px at full resolution, covers the residual misalignment of most scenes.

Pixels without a reliable correspondence account for 5.2\% of the image in the median scene and 9.6\% on average. The higher mean indicates a skewed distribution, in which a minority of scenes with large foreground objects close to the cameras contain wide occluded regions. In these regions no correct RGB correspondence exists, which is the situation the gate of TANGO is designed to handle.

\begin{table}[!ht]
\centering
\caption{Misalignment statistics on DUET (403 scenes). The residual is the displacement not explained by the best-fitting homography, and P95 denotes its 95th percentile within a scene.}
\label{tab:supp_misalignment}
\small
\setlength{\tabcolsep}{4pt}
\begin{tabular}{l c}
\toprule
Statistic & Value \\
\midrule
P95 residual, median scene [px] & 11.1 \\
P95 residual, 90th-percentile scene [px] & 26.4 \\
Scenes with P95 residual $>$ 4\,px [\%] & 72.7 \\
Pixels with residual $>$ 4\,px, median scene [\%] & 22.0 \\
Pixels without correspondence, median / mean [\%] & 5.2 / 9.6 \\
\bottomrule
\end{tabular}
\end{table}
 
\begin{figure*}[t]
    \centering
    \includegraphics[width=\linewidth]{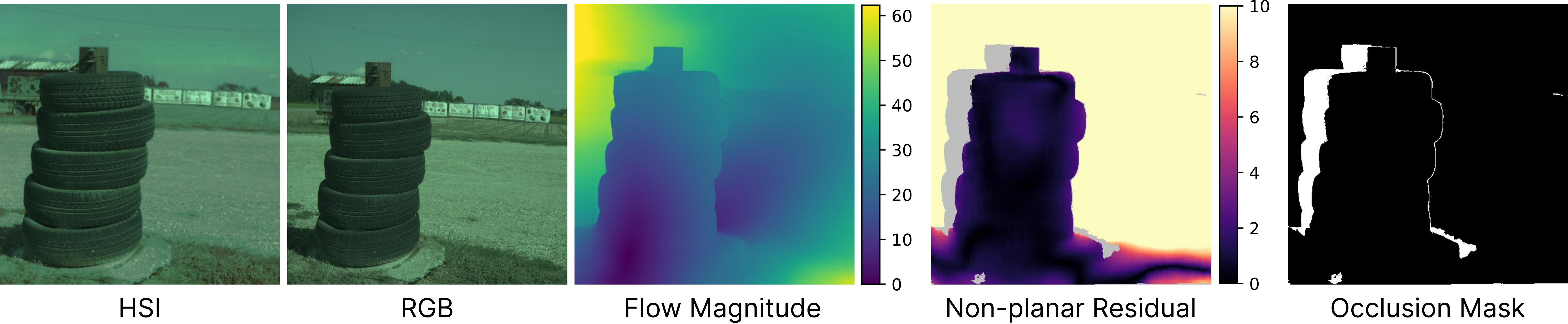}
    \caption{Correspondence analysis on a DUET scene. From left to right: hyperspectral pseudo-RGB, RGB image, flow magnitude in pixels, non-planar residual with respect to the best-fitting homography, clipped at 10\,px, and occlusion mask, where white pixels fail the forward--backward consistency check or fall outside the RGB view and are shown in gray in the residual map. Displacements are measured after the field-of-view matching and are therefore relative to the roughly aligned foreground.}
    \label{fig:supp_corr_maps}
\end{figure*}

\Cref{fig:supp_corr_maps} shows the correspondence analysis on a representative scene with a foreground object in front of a distant background. Since the field-of-view matching is estimated mostly from correspondences on textured foreground objects, the foreground is already roughly aligned, with a flow magnitude of about 10--20\,px, whereas the background is displaced by 30--60\,px. The non-planar residual is therefore close to zero on the foreground, which is the layer fitted by RANSAC, and exceeds 10\,px on most of the background, showing that no single homography can register both depth layers. The occlusion mask highlights a strip on the left side of the foreground object, where the background visible from the hyperspectral camera is hidden in the RGB view because of the horizontal baseline. These are the regions where the RGB guidance cannot be trusted, of the same kind as the pixels without reliable correspondence counted in \cref{tab:supp_misalignment}.

\section{Additional Results}
\label{sec:supp_results}

This section complements the experimental evaluation of the main paper. \Cref{sec:supp_qualitative} shows the reconstructed images for the DUET scenes whose error maps are reported in the main paper, allowing a direct visual comparison of the recovered spatial details. \Cref{sec:supp_tradeoff} analyzes the trade-off between accuracy and computational cost. \Cref{sec:supp_real} reports the results on REAL at all scale factors, extending the comparison of the main paper, which is limited to $\times8$, to $\times4$ and $\times16$. 

\subsection{Additional Qualitative Results}
\label{sec:supp_qualitative}

\begin{figure*}[t]
    \centering
    \includegraphics[width=\linewidth]{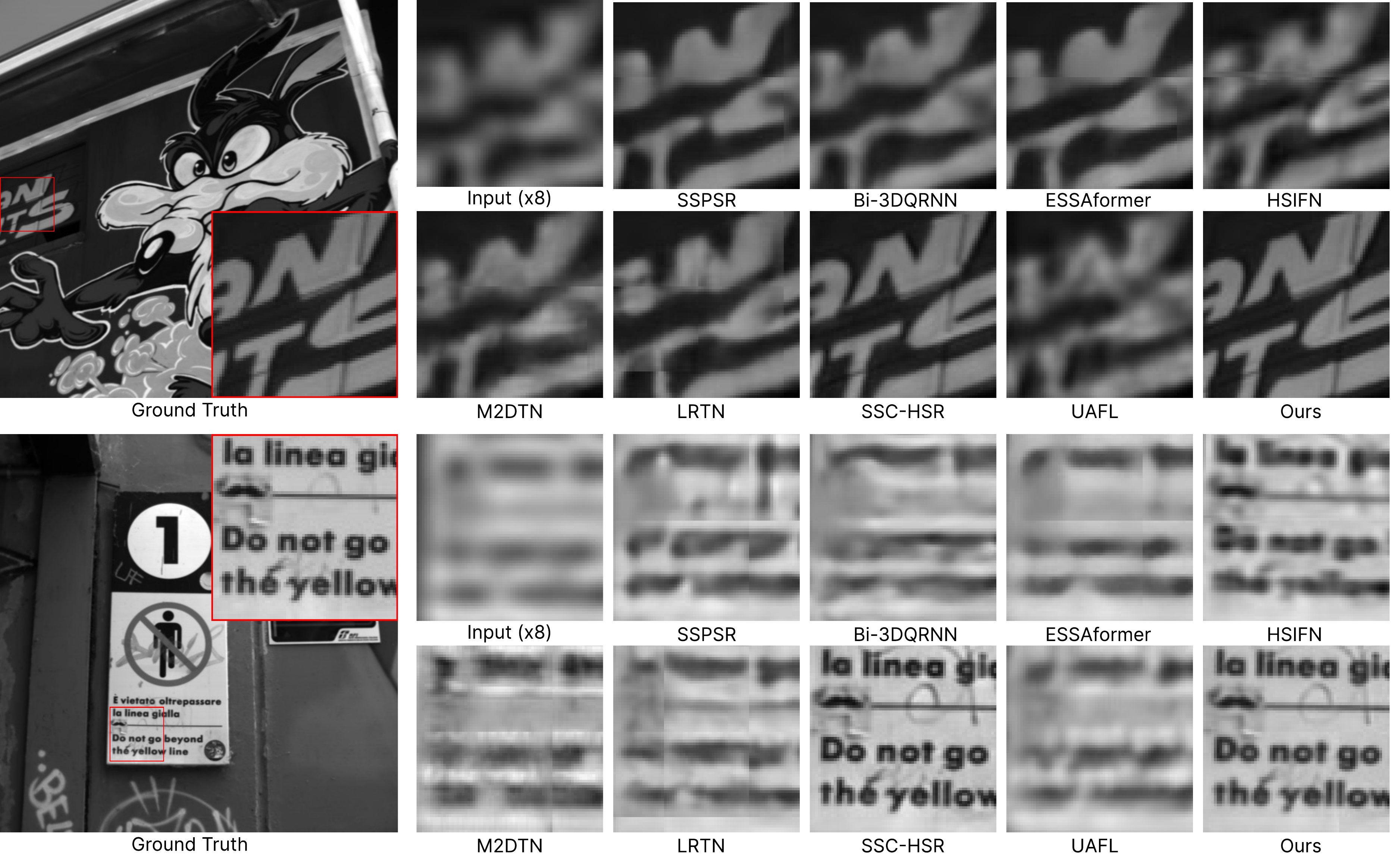}
    \caption{Reconstructions of band 20 (600\,nm) for the DUET scenes of \cref{fig:DUET_qualitative} at $\times8$. The red boxes indicate the enlarged regions. TANGO recovers sharp spatial details and preserves object boundaries, while methods without explicit alignment show blurring or artifacts caused by misaligned RGB guidance.}
    \label{fig:supp_qual_DUET}
\end{figure*}

\Cref{fig:supp_qual_DUET} complements the error maps of \cref{fig:DUET_qualitative} in the main paper by showing the reconstructed band for the same two scenes, so that the differences between methods can be assessed directly on the images rather than through the error magnitude. Both scenes contain fine, high-contrast structures, namely painted letters and printed text, which are lost in the $\times8$ low-resolution input and can only be recovered by transferring spatial detail from the RGB guidance.

On the mural, single-image methods produce smooth reconstructions in which the contours of the letters remain blurred, since the missing high frequencies cannot be inferred from the low-resolution input alone. RGB-guided methods recover sharper edges, but several of them show ghosting, doubled contours, or visible tile seams, which arise when RGB details are injected at wrong positions because of the residual cross-camera misalignment. TANGO restores the contours of the letters with fewer such artifacts, consistently with its lower reconstruction error in this region.

On the sign, the text is unreadable in the reconstructions of single-image methods and remains blurred for most RGB-guided methods. Only TANGO and SSC-HSR reconstruct legible characters, as both explicitly align the RGB guidance to the hyperspectral view before fusion. Between the two, TANGO produces cleaner strokes and fewer residual artifacts around the characters, in line with the error maps of the main paper. Overall, these examples show that the quantitative gains of TANGO correspond to visible improvements in the recovery of fine structures, which are the regions most affected by cross-camera misalignment.

\subsection{Accuracy--Complexity Trade-off}
\label{sec:supp_tradeoff}

\begin{figure}[t]
    \centering
    \includegraphics[width=\columnwidth]{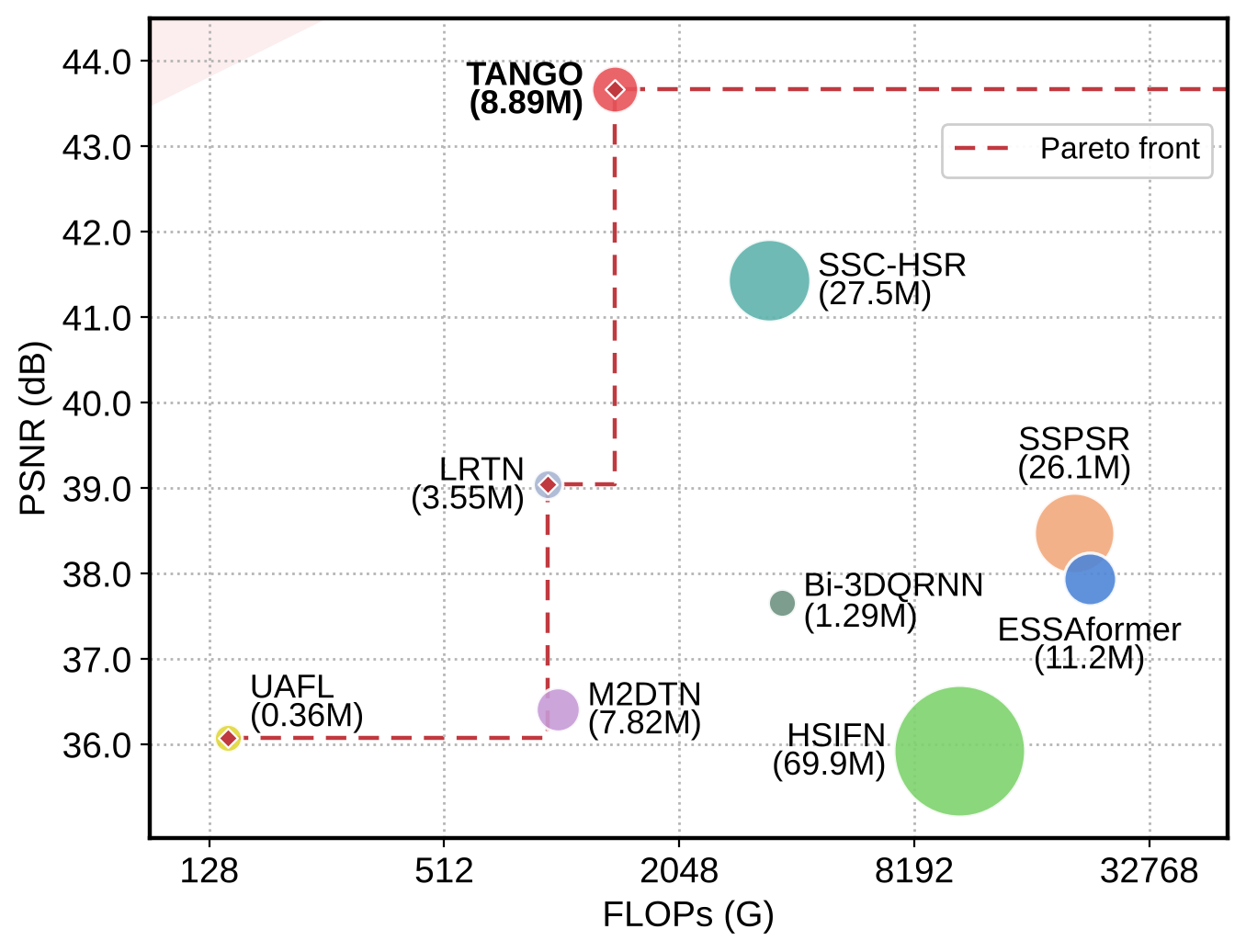}
    \caption{PSNR versus FLOPs on DUET at $\times4$. The bubble area is proportional to the number of parameters, reported in brackets. The dashed line is the Pareto front with respect to FLOPs and PSNR, and better trade-offs lie toward the upper left, highlighted by the shaded corner. FLOPs are computed on $512\times512$ images.}
    \label{fig:supp_pareto}
\end{figure}

\Cref{fig:supp_pareto} compares all methods on DUET at $\times4$ in terms of PSNR and FLOPs, with the bubble area proportional to the number of parameters. The dashed line is the Pareto front with respect to FLOPs and PSNR: a method lies on the front if no other method reaches a higher PSNR with fewer FLOPs. Only three methods lie on the front, namely UAFL, LRTN, and TANGO, and TANGO is the only one above 39\,dB.

TANGO dominates five of the eight competing methods, since SSC-HSR, Bi-3DQRNN, HSIFN, SSPSR, and ESSAformer all require more FLOPs and reach a lower PSNR. The closest competitor, SSC-HSR, is 2.24\,dB below TANGO while requiring 2.5$\times$ more FLOPs and 3.1$\times$ more parameters. Single-image methods such as SSPSR and ESSAformer are the most expensive, with about 15$\times$ more FLOPs than TANGO, and remain more than 5\,dB below it. Moving along the front from LRTN to TANGO, the alignment stages increase the cost by 1.5$\times$ in FLOPs and improve PSNR by 4.63\,dB, which is by far the largest step of the front. The remaining methods on the front, UAFL and LRTN, are cheaper but reach much lower accuracy, so TANGO provides the best accuracy among all methods at a moderate computational cost.

\subsection{Complete Results on REAL}
\label{sec:supp_real}

All methods are retrained on REAL with the official split of 47 training and 10 test pairs, using the same degradation, normalization, and evaluation protocol as for DUET; the results may therefore differ from those originally reported. \Cref{tab:supp_real} extends \cref{tab:real_results} of the main paper to $\times4$ and $\times16$.

\begin{table*}[t]
\centering
\caption{Quantitative comparison on REAL at different scale factors. PSNR is reported in dB. Model complexity is reported for the reference $\times4$ configuration on REAL images. $\uparrow$ indicates that higher values are better, while $\downarrow$ indicates that lower values are better. The best and second-best results are highlighted in \textbf{bold} and \underline{underline}, respectively.}
\label{tab:supp_real}

\small
\setlength{\tabcolsep}{4pt}

\begin{tabular}{l ccc ccc ccc ll}
\toprule

\multirow{2}{*}{\textbf{Method}} &
\multicolumn{3}{c}{\textbf{Scale Factor} $\times4$} &
\multicolumn{3}{c}{\textbf{Scale Factor} $\times8$} &
\multicolumn{3}{c}{\textbf{Scale Factor} $\times16$} &
\multirow{2}{*}{\shortstack{\textbf{Params}\\\textbf{[M]}}} &
\multirow{2}{*}{\shortstack{\textbf{FLOPs}\\\textbf{[G]}}} \\

\cmidrule(lr){2-4}
\cmidrule(lr){5-7}
\cmidrule(lr){8-10}

&
PSNR$\uparrow$ & SSIM$\uparrow$ & SAM$\downarrow$ &
PSNR$\uparrow$ & SSIM$\uparrow$ & SAM$\downarrow$ &
PSNR$\uparrow$ & SSIM$\uparrow$ & SAM$\downarrow$ &
& \\

\midrule

Bicubic
& 33.678 & 0.8868 & 0.04003
& 28.675 & 0.7810 & 0.05743
& 24.848 & 0.7024 & 0.08372
& \multicolumn{1}{c}{--} & \multicolumn{1}{c}{--} \\

SSPSR~\cite{FPB2020}
& 39.521 & 0.9498 & \textbf{0.03468}
& 32.404 & 0.8496 & 0.04912
& 27.317 & 0.7517 & 0.06652
& 26.08 & 21066.1 \\

Bi-3DQRNN~\cite{9351612}
& 38.366 & 0.9398 & 0.04044
& 31.926 & 0.8379 & 0.05433
& 27.078 & 0.7414 & 0.08212
& \underline{1.288} & 4100.9 \\

ESSAformer~\cite{EssaFormer}
& 37.906 & 0.9368 & 0.04445
& 31.443 & 0.8304 & 0.06384
& 26.942 & 0.7379 & 0.08295
& 11.194 & 21795.5 \\

\midrule

HSIFN~\cite{lai2024hyperspectral}
& 36.928 & 0.9321 & 0.05550
& 33.181 & 0.8795 & 0.07013
& 29.161 & 0.8317 & 0.09586
& 69.922 & 14276.4 \\

M2DTN~\cite{Dong_Xu_Qu_Hou_2024}
& 35.921 & 0.9373 & 0.06850
& 34.027 & 0.9121 & 0.06966
& 30.591 & 0.8825 & 0.08218
& 7.823 & \underline{1001.0} \\

LRTN~\cite{liu2025low}
& \underline{40.667} & 0.9600 & 0.03752
& \underline{36.424} & \underline{0.9236} & \underline{0.04909}
& \underline{32.754} & \underline{0.8972} & \underline{0.06163}
& 3.551 & 1032.6 \\

SSC-HSR~\cite{zhang2025unaligned}
& 40.392 & \underline{0.9609} & 0.04092
& 35.532 & 0.9221 & 0.05945
& 31.210 & 0.8816 & 0.08569
& 27.451 & 3924.3 \\

UAFL~\cite{zhang2026enhancing}
& 39.077 & 0.9487 & 0.03966
& 34.564 & 0.9038 & 0.05593
& 29.944 & 0.8594 & 0.08679
& \textbf{0.358} & \textbf{156.5} \\

\textbf{TANGO}
& \textbf{41.558} & \textbf{0.9671} & \underline{0.03546}
& \textbf{37.368} & \textbf{0.9410} & \textbf{0.04537}
& \textbf{32.775} & \textbf{0.8998} & \textbf{0.05507}
& 8.887 & 1534.5 \\

\bottomrule
\end{tabular}
\end{table*}

TANGO achieves the best PSNR and SSIM at all scale factors and the best SAM at $\times8$ and $\times16$, while SSPSR obtains a slightly lower SAM at $\times4$. The strongest competitor on REAL is LRTN, which TANGO outperforms by 0.89\,dB at $\times4$ and 0.94\,dB at $\times8$. At $\times16$ the PSNR gap reduces to 0.02\,dB, but TANGO still reduces SAM by 11\%, from 0.06163 to 0.05507. With respect to LRTN, the SAM reduction grows with the scale factor, from 5\% at $\times4$ to 8\% at $\times8$ and 11\% at $\times16$, indicating that the alignment stages mainly improve spectral fidelity when the reconstruction relies more heavily on the RGB guidance.

The role of the RGB guidance also emerges from the comparison with single-image methods. At $\times4$, the best single-image method, SSPSR, is only 2.04\,dB below TANGO and outperforms several RGB-guided methods, since the low-resolution input still contains most of the spatial information. As the scale factor increases, the gap between TANGO and the best single-image method widens to 4.96\,dB at $\times8$ and 5.46\,dB at $\times16$, and all RGB-guided methods outperform single-image ones, confirming that the guidance becomes essential under severe spatial degradation.

In terms of complexity, TANGO requires 8.887\,M parameters and 1534.5\,GFLOPs on REAL images, about 2.6$\times$ fewer FLOPs than SSC-HSR, while remaining heavier than LRTN because of the alignment stages. Compared with DUET, the margins over methods without explicit alignment are smaller, consistently with the different nature of the misalignment in REAL, where both views are acquired with hyperspectral cameras and the RGB guidance is synthesized, so cross-sensor discrepancies are absent.
 
\section{Complete Ablation Study}
\label{sec:supp_ablation}
 
This section extends the ablation of the main paper to all scale factors and analyzes where the local alignment and the gate are beneficial. All variants are trained with the same recipe as TANGO.

\subsection{Global and Local Alignment}
\label{sec:supp_ablation_align}

\Cref{tab:supp_ablation} extends \cref{tab:tango_ablation} of the main paper to $\times4$, $\times8$, and $\times16$. The trends observed at $\times8$ hold at the other scale factors. Without global alignment, local alignment alone improves PSNR by 0.40\,dB at $\times4$, 0.35\,dB at $\times8$, and 0.72\,dB at $\times16$, with only 0.079\,M additional parameters, showing that feature matching can partially compensate for the cross-camera misalignment even without an explicit flow. The largest gains come from the end-to-end optimized global alignment, on top of which local alignment still adds 0.35\,dB at $\times4$, 0.23\,dB at $\times8$, and 0.56\,dB at $\times16$. The contribution of local alignment is therefore largest at $\times16$, where the low-resolution input carries the least spatial information and the reconstruction depends most on the RGB guidance. Local alignment also reduces SAM in all nine configurations of \cref{tab:supp_ablation}, and with respect to the global-only configuration it lowers SAM by about 5\% at every scale, indicating that more accurate correspondences improve spectral fidelity as well as spatial detail.

The frozen pretrained RAFT behaves differently across scales. At $\times4$ and $\times8$ it yields gains of 3.52\,dB and 2.05\,dB over the configuration without global alignment, whereas at $\times16$ it degrades the reconstruction by 3.47\,dB. At this scale, the bicubic pseudo-RGB obtained from a $32\times32$ input is too blurry for a generic flow estimator, which produces wrong warps that the local alignment cannot compensate: on top of the frozen RAFT, local alignment leaves PSNR almost unchanged and only slightly reduces SAM. End-to-end fine-tuning removes this failure and reaches 31.863\,dB, confirming that adapting the global correspondence estimator to the task is essential, especially under severe spatial degradation.

\begin{table*}[t]
\centering
\caption{Complete ablation study on DUET. PSNR is reported in dB. $\times$ denotes a disabled component, \faSnowflake~denotes a frozen pretrained RAFT, and \checkmark~denotes an enabled component, with end-to-end optimization for the global alignment. $\uparrow$ indicates that higher values are better, while $\downarrow$ indicates that lower values are better. The best and second-best results are highlighted in \textbf{bold} and \underline{underline}, respectively.}
\label{tab:supp_ablation}

\small
\setlength{\tabcolsep}{4pt}

\begin{tabular}{cc ccc ccc ccc c}
\toprule

\multirow{2}{*}{\textbf{Global}} &
\multirow{2}{*}{\textbf{Local}} &
\multicolumn{3}{c}{\textbf{Scale Factor} $\times4$} &
\multicolumn{3}{c}{\textbf{Scale Factor} $\times8$} &
\multicolumn{3}{c}{\textbf{Scale Factor} $\times16$} &
\multirow{2}{*}{\shortstack{\textbf{Params}\\\textbf{[M]}}} \\

\cmidrule(lr){3-5}
\cmidrule(lr){6-8}
\cmidrule(lr){9-11}

& &
PSNR$\uparrow$ & SSIM$\uparrow$ & SAM$\downarrow$ &
PSNR$\uparrow$ & SSIM$\uparrow$ & SAM$\downarrow$ &
PSNR$\uparrow$ & SSIM$\uparrow$ & SAM$\downarrow$ &
\\

\midrule

\multirow{2}{*}{$\times$}
& $\times$
& 39.046 & 0.9597 & 0.03127
& 33.292 & 0.9004 & 0.05001
& 29.058 & 0.8329 & 0.07061
& 3.551 \\
& \checkmark
& 39.449 & 0.9624 & 0.03027
& 33.644 & 0.9059 & 0.04914
& 29.774 & 0.8485 & 0.06819
& 3.630 \\

\midrule

\multirow{2}{*}{\faSnowflake}
& $\times$
& 42.570 & 0.9807 & 0.02653
& 35.339 & 0.9330 & 0.04221
& 25.587 & 0.7250 & 0.08254
& 3.551 \\
& \checkmark
& 42.739 & 0.9813 & 0.02624
& 35.676 & 0.9367 & 0.04060
& 25.517 & 0.7239 & 0.08206
& 3.630 \\

\midrule

\multirow{2}{*}{\checkmark}
& $\times$
& \underline{43.325} & \underline{0.9824} & \underline{0.02593}
& \underline{37.888} & \underline{0.9534} & \underline{0.03942}
& \underline{31.863} & \underline{0.8967} & \underline{0.06075}
& 8.809 \\
& \checkmark
& \textbf{43.672} & \textbf{0.9833} & \textbf{0.02467}
& \textbf{38.116} & \textbf{0.9560} & \textbf{0.03762}
& \textbf{32.418} & \textbf{0.9047} & \textbf{0.05725}
& 8.887 \\

\bottomrule
\end{tabular}
\end{table*}

\subsection{Analysis of the Gate}
\label{sec:supp_gate}

\noindent\textbf{Effect across scale factors.}
\Cref{tab:supp_gate_ablation} extends \cref{tab:gate_ablation} of the main paper to all scale factors. The gate adds only 0.063\,M parameters and improves all metrics at every scale, with PSNR gains of 0.42\,dB at $\times4$, 0.25\,dB at $\times8$, and 0.28\,dB at $\times16$. At $\times4$ and $\times8$, local alignment without the gate does not improve PSNR over the global-only configuration and only slightly reduces SAM: injecting the attended RGB features without modulation offsets the benefit of the refined correspondences. At $\times16$, local alignment improves the reconstruction by 0.28\,dB even without the gate, since any additional RGB detail is valuable when the low-resolution input is so limited, and the gate doubles this gain.

\begin{table*}[t]
\centering
\caption{Effect of the gate in the local alignment on DUET at different scale factors. PSNR is reported in dB. Both variants use end-to-end optimized global alignment.}
\label{tab:supp_gate_ablation}

\small
\setlength{\tabcolsep}{4pt}

\begin{tabular}{l ccc ccc ccc l}
\toprule

\multirow{2}{*}{\textbf{Local alignment}} &
\multicolumn{3}{c}{\textbf{Scale Factor} $\times4$} &
\multicolumn{3}{c}{\textbf{Scale Factor} $\times8$} &
\multicolumn{3}{c}{\textbf{Scale Factor} $\times16$} &
\multirow{2}{*}{\shortstack{\textbf{Params}\\\textbf{[M]}}} \\

\cmidrule(lr){2-4}
\cmidrule(lr){5-7}
\cmidrule(lr){8-10}

&
PSNR$\uparrow$ & SSIM$\uparrow$ & SAM$\downarrow$ &
PSNR$\uparrow$ & SSIM$\uparrow$ & SAM$\downarrow$ &
PSNR$\uparrow$ & SSIM$\uparrow$ & SAM$\downarrow$ &
\\

\midrule

w/o gate
& 43.249 & 0.9820 & 0.02577
& 37.867 & 0.9528 & 0.03886
& 32.139 & 0.8987 & 0.06025
& 8.824 \\

w/ gate
& \textbf{43.672} & \textbf{0.9833} & \textbf{0.02467}
& \textbf{38.116} & \textbf{0.9560} & \textbf{0.03762}
& \textbf{32.418} & \textbf{0.9047} & \textbf{0.05725}
& 8.887 \\

\bottomrule
\end{tabular}
\end{table*}

\noindent\textbf{Per-region error.}
To analyze where the gate and the local alignment are beneficial, we partition the test pixels of DUET according to the misalignment analysis of \cref{sec:supp_misalignment}: planar pixels, whose residual with respect to the homography is below 1\,px; pixels with a residual of 1--4\,px and above 4\,px; and pixels without a reliable correspondence. \Cref{tab:supp_gate_regions} reports the per-pixel mean absolute error at $\times8$ for the global-only configuration, TANGO without the gate, and the complete TANGO. Removing the gate increases the error in all regions, and on pixels with residual parallax the local alignment without the gate is even worse than the global-only configuration, suggesting that unmodulated attention can transfer RGB details from wrong matches where correspondences are ambiguous. The largest improvement of TANGO over the global-only configuration is obtained on pixels without a reliable correspondence, with a 6.2\% lower error, where the global flow is least reliable. Pixels with large but consistently estimated parallax, instead, are already handled by the global alignment, and TANGO improves them by only 0.5\%. Global and local alignment therefore address different failure modes: the former compensates for consistent displacements, including depth-dependent parallax, while the latter, modulated by the gate, mainly recovers the regions where the global correspondence fails.

\begin{table*}[t]
\centering
\caption{Per-pixel MAE ($\times10^{-3}$) on the DUET test set at $\times8$, by region. The last column reports the relative change of TANGO with respect to the global-only configuration.}
\label{tab:supp_gate_regions}
\small
\setlength{\tabcolsep}{3pt}
\begin{tabular}{l c ccc c}
\toprule
Region & Pixels [\%] & Global & w/o gate & TANGO & $\Delta$ \\
\midrule
Planar ($<$1\,px)      & 51.6 & 6.04  & 6.01  & \textbf{5.73}  & $-5.1\%$ \\
1--4\,px               & 22.2 & 7.73  & 7.87  & \textbf{7.50}  & $-3.1\%$ \\
$>$4\,px               & 15.8 & 9.87  & 10.07 & \textbf{9.81}  & $-0.5\%$ \\
No correspondence      & 10.4 & 12.28 & 12.06 & \textbf{11.53} & $-6.2\%$ \\
\bottomrule
\end{tabular}
\end{table*}
 
\noindent\textbf{Gate values.}
The average gate value increases from coarse to fine scales, from 0.35 at quarter resolution to 0.40 at half and 0.65 at full resolution, so that the RGB features are attenuated more strongly at the coarser scales, where the wider search window makes correspondences more ambiguous. At full resolution, 86\% of the gate values lie between 0.5 and 0.7, and the gate is not a binary occlusion mask: its average value is similar on pixels with and without a reliable correspondence. The gate therefore acts as a learned, mostly smooth modulation of the RGB guidance rather than as an explicit occlusion detector, and its effect is nevertheless largest where correspondences fail, as shown in \cref{tab:supp_gate_regions}.

\end{document}